\documentclass[11pt]{article}

\usepackage[margin=1in]{geometry}

\usepackage[utf8]{inputenc} 
\usepackage[T1]{fontenc}    
\usepackage{microtype}
\usepackage{graphicx}
\usepackage{multicol}
\usepackage{booktabs} 
\usepackage{multirow}
\usepackage{amssymb,amsmath,amsthm,amsfonts}
\usepackage{tikz}
\usetikzlibrary{calc}
\usepackage{bm}
\usepackage{textgreek}
\usepackage{mathtools}
\usepackage{enumitem}
\usepackage{authblk}
\usepackage{graphicx}
\usepackage[font=small]{caption}
\usepackage[linesnumbered,ruled,vlined,boxed,algo2e]{algorithm2e}
\usepackage{physics}
\usepackage{footnote}
\usepackage{xcolor}
\usepackage{mathrsfs}
\usepackage{bbm}
\usepackage{makecell}
\usepackage[ruled,vlined,linesnumbered]{algorithm2e}
\usepackage{algorithmic}

\usepackage[colorlinks]{hyperref}

\usepackage{verbatim}

\usepackage[numbers,sort]{natbib}

\usepackage{subcaption}

\newtheorem{theorem}{Theorem}

\newtheorem{lemma}{Lemma}
\newtheorem{proposition}{Proposition}

\newtheorem{problem}{Problem}

\newcommand{\eq}[1]{(\ref{eq:#1})}

\newcommand{\thm}[1]{\hyperref[thm:#1]{Theorem~\ref*{thm:#1}}}
\newcommand{\cor}[1]{\hyperref[cor:#1]{Corollary~\ref*{cor:#1}}}
\newcommand{\defn}[1]{\hyperref[defn:#1]{Definition~\ref*{defn:#1}}}
\newcommand{\lem}[1]{\hyperref[lem:#1]{Lemma~\ref*{lem:#1}}}
\newcommand{\prop}[1]{\hyperref[prop:#1]{Proposition~\ref*{prop:#1}}}
\newcommand{\assum}[1]{\hyperref[assum:#1]{Assumption~\ref*{assum:#1}}}
\newcommand{\fig}[1]{\hyperref[fig:#1]{Figure~\ref*{fig:#1}}}
\newcommand{\tab}[1]{\hyperref[tab:#1]{Table~\ref*{tab:#1}}}
\newcommand{\algo}[1]{\hyperref[algo:#1]{Algorithm~\ref*{algo:#1}}}
\renewcommand{\sec}[1]{\hyperref[sec:#1]{Section~\ref*{sec:#1}}}
\newcommand{\append}[1]{\hyperref[append:#1]{Appendix~\ref*{append:#1}}}
\newcommand{\fac}[1]{\hyperref[fac:#1]{Fact~\ref*{fac:#1}}}
\newcommand{\lin}[1]{\hyperref[lin:#1]{Line~\ref*{lin:#1}}}
\newcommand{\prob}[1]{\hyperref[prob:#1]{Problem~\ref*{prob:#1}}}

\def\>{\rangle}
\def\<{\langle}

\newcommand{\vect}[1]{\ensuremath{\mathbf{#1}}}
\newcommand{\x}{\ensuremath{\mathbf{x}}}
\newcommand{\y}{\ensuremath{\mathbf{y}}}
\newcommand{\z}{\ensuremath{\mathbf{z}}}
\newcommand{\g}{\ensuremath{\mathbf{g}}}
\newcommand{\e}{\ensuremath{\mathbf{e}}}
\newcommand{\w}{\ensuremath{\mathbf{w}}}

\newcommand{\Z}{\mathbb{Z}}
\newcommand{\R}{\mathbb{R}}

\newcommand{\E}{\mathbb{E}}

\DeclareMathOperator{\diag}{diag}

\DeclareMathOperator{\comp}{Comp}

\DeclareMathOperator{\sgn}{sgn}

\renewcommand{\d}{\mathrm{d}}
\renewcommand{\x}{\vect{x}}
\renewcommand{\u}{\vect{u}}
\renewcommand{\v}{\vect{v}}

\renewcommand{\emptyset}{\varnothing}
\def\:{\hbox{\bf:}}

\SetKwInput{KwInput}{Input} 
\SetKwInput{KwParameter}{Parameters}           
\SetKwInput{KwOutput}{Output}              
\SetKwInput{KwReturn}{Return}
\SetKwComment{Comment}{/* }{ */}
\SetKwBlock{RepeatBlock}{repeat}{end repeat}

\let\oldnl\nl
\newcommand{\nonl}{\renewcommand{\nl}{\let\nl\oldnl}}

\usepackage[textsize=tiny]{todonotes}

\newcommand\blfootnote[1]{%
  \begingroup
  \renewcommand\thefootnote{}\footnote{#1}%
  \addtocounter{footnote}{-1}%
  \endgroup
}

\title{Gradient Testing and Estimation by Comparisons}

\date{}

\author{
  Xiwen Tao$^{1,3,}$\thanks{Equal contribution. } \qquad Chenyi Zhang$^{2,}$\protect\footnotemark[1]\qquad Helin Wang$^{1,3}$\qquad Yexin Zhang$^{3,4}$\qquad Tongyang Li$^{3,4,}$\thanks{Corresponding author. Email: tongyangli@pku.edu.cn} \\
  $^1$ School of Electronics Engineering and Computer Science, Peking University\\
  $^2$ Computer Science Department, Stanford University \\
  $^3$ Center on Frontiers of Computing Studies, Peking University\\
  $^4$ School of Computer Science, Peking University\\
}

\begin{document}

\maketitle

\begin{abstract}
We study gradient testing and gradient estimation of smooth functions using only a comparison oracle that, given two points, indicates which one has the larger function value. For any smooth $f\colon\mathbb R^n\to\mathbb R$, $\x\in\R^n$, and $\varepsilon>0$, we design a gradient testing algorithm that determines whether the normalized gradient $\nabla f(\x)/\|\nabla f(\x)\|$ is $\varepsilon$-close or $2\varepsilon$-far from a given unit vector $\v$ using $O(1)$ queries, as well as a gradient estimation algorithm that outputs an $\varepsilon$-estimate of $\nabla f(\x)/\|\nabla f(\x)\|$ using $O(n\log(1/\varepsilon))$ queries which we prove to be optimal. Furthermore, we study gradient estimation in the quantum comparison oracle model where queries can be made in superpositions, and develop a quantum algorithm using $O(\log (n/\varepsilon))$ queries. \blfootnote{This work subsumes the note ``Comparisons are all You need for optimizing smooth functions''~\cite{zhang2024comparisons}.} 
\end{abstract}

\section{Introduction}
Optimization is foundational to machine learning, since the training of a neural network is equivalent to the minimization of its loss function. However, there are common scenarios where access to gradients is infeasible or computationally prohibitive, such as black-box adversarial attack on neural networks~\cite{papernot2017practical,madry2018towards,chen2017zoo} and policy search in reinforcement learning~\cite{salimans2017evolution,choromanski2018structured}. 
This motivates the study of optimization algorithms that operate with zeroth-order information, i.e., function-value evaluations. A series of work has established rigorous convergence and complexity guarantees for such methods in both convex optimization~\cite{duchi2015optimal,nesterov2017random} and nonconvex optimization~\cite{ghadimi2013stochastic,fang2018spider,jin2018local,ji2019improved,zhang2022zeroth,vlatakis2019efficiently,balasubramanian2022zeroth}.

More recently, optimization algorithms are soliciting for even less information. In particular, an intriguing setting is to only have access to comparisons of function values. Formally, for a function $f\colon\R^n\to\R$, its comparison oracle is defined as $O_f^{\comp}\colon\R^n\times\R^n\rightarrow\{-1,1\}$ that for any pair of inputs $(\x,\y)\in\R^n\times\R^n$, it satisfies
\begin{align}\label{eq:comparison}
O_f^{\comp}=
    \begin{cases}
    1 & f(\x)\ge f(\y) \\
    -1 & f(\x)\le f(\y)
    \end{cases}.
\end{align}
(When $f(\x)=f(\y)$, it is allowed to output either $1$ or $-1$.)

Several studies have developed optimization algorithms under this comparison setting. 
As a class of gradient-free optimization methods (see the survey by~\cite{larson2019derivative}), these works used comparisons once or multiple times per iteration to approximately estimate the direction of the gradient. This idea can be traced back to~\cite{nelder1965simplex}, which iteratively updates a simplex of points based on comparisons to seek lower objective values. However, the Nelder-Mead method may fail to converge to a stationary point for smooth functions~\cite{dennis1991direct}. Recent works, including the Stochastic Three-Point (STP) method~\cite{bergou2020stochastic} and dueling optimization~\cite{saha2021dueling}, update the optimization direction by randomly sampling directions per iteration and using comparisons to determine the relative ordering of function values. 

Furthermore, Ref.~\cite{cheng2019sign} estimated a descent direction using comparisons between candidate points based on randomized probing at each iteration. Ref.~\cite{karabag2021smooth} showed how to use comparison queries or direction preference queries inside ellipsoid methods for solving smooth convex problems. In a different line of work,~\cite{cai2022one} leveraged ideas from one-bit compressed sensing to recover gradient information from noisy comparison feedback. More recently, Ref.~\cite{tang2023zeroth} introduced a rank-based stochastic gradient estimator that uses the partial ordering of the best candidates to infer a descent direction. 

Despite the importance of gradients in optimization, prior work has not pinned down the optimal algorithm for gradient estimation using comparison queries. An existing result is due to~\cite{karabag2021smooth}, which achieved $O(n^2 \log n)$ comparison queries per gradient direction estimate at a fixed angular accuracy. Another related work is dueling optimization~\cite{saha2021dueling}, which provided convergence guarantees based on update directions that are aligned with the true gradient in expectation; however, their analysis does not characterize the query complexity required to estimate gradient directions to a designated accuracy. 

Beyond that, gradient testing is also a natural problem to explore. In general, property testing decides whether a given object has a certain property or is significantly different from any object that has the property. It typically brings more efficient algorithms than property estimation, and may circumvent significant costs when the input size is large or an exact algorithm is expensive. Therefore, property testing is a main algorithmic topic~\cite{ron2010algorithmic,goldreich2017introduction}, especially on discrete problems. However, its study on continuous optimization is scarce, which motivates us to study gradient testing together with gradient estimation.

Recently, following the rapid advancement of quantum computing~\cite{preskill2018quantum,preskill2025beyond}, quantum algorithms that provide speedup for solving optimization problems have been systematically investigated. These include constrained optimization problems covering linear programs~\cite{casares2020quantum,bouland2023quantum,gao2023logarithmic}, second-order cone programming~\cite{kerenidis2021quantum,garrido2025quantum}, and semidefinite programs~\cite{brandao2017quantum,brandao2017SDP,van2019improvements,van2020quantum}. General convex optimization~\cite{chakrabarti2020optimization,van2020convex,sidford2023quantum} and nonconvex optimization~\cite{zhang2021quantum,liu2022quantum,leng2023quantum,chen2023quantum,leng2025subexponential} are also extensively studied. See the survey by~\cite{dalzell2023quantum} for more details. In the quantum setting, it is also very natural to study optimization by comparisons, particularly because current quantum computers are noisy and simpler information directly benefits implementation of quantum algorithms in the near term. In particular, we study the setting where we are given a quantum comparison oracle
\begin{align}\label{eq:comparison-oracle-quantum}
    O_{f,Q}^{\mathrm{Comp}}\ket{\x}\ket{\y}\ket{z}\to\ket{\x}\ket{\y}\ket*{z\oplus O_f^{\mathrm{Comp}}(\x,\y)}
\end{align}
which is the quantum generalization of the comparison oracle $O_f^{\mathrm{Comp}}$ in \eqref{eq:comparison}, where $\oplus$ means XOR.  

\paragraph{Main results.}
In this paper, we systematically study gradient testing and estimation, formally stated as follows:
\begin{problem}[Gradient testing]\label{prob:testing}
Let $f\colon\mathbb R^n\to\mathbb R$ be $L$-smooth, $\varepsilon\in(0,1/\sqrt{2})$ and $\gamma>0$ be fixed parameters. Given a point $\x\in\mathbb R^n$ and a unit vector $\v\in\mathbb R^n$, the goal is to decide whether
\[
\Big\|\frac{\nabla f(\x)}{\|\nabla f(\x)\|}-\v\Big\|\le \varepsilon
\quad\text{or}\quad
\Big\|\frac{\nabla f(\x)}{\|\nabla f(\x)\|}-\v\Big\|>2\varepsilon,
\]
where we are promised that one of the two cases holds and $\|\nabla f(\x)\|\ge \gamma$.
\end{problem}

\begin{problem}[Gradient estimation]\label{prob:estimation}
Let $f\colon\mathbb R^n\to\mathbb R$ be $L$-smooth, $\varepsilon\in(0,1/\sqrt{2})$ and $\gamma>0$ be fixed parameters. Given a point $\x\in\mathbb R^n$ satisfying $\|\nabla f(\x)\|\ge \gamma$, the goal is to output a unit vector $\v\in\mathbb R^n$ satisfying $\big\|\frac{\nabla f(\x)}{\|\nabla f(\x)\|}-\v\big\|\le \varepsilon$.
\end{problem}

Both problems are scale-invariant: multiplying the objective by a positive constant does not change either the comparison oracle or the normalized gradient direction. The lower bound $\|\nabla f(\x)\|\ge \gamma$ is therefore a nondegeneracy condition rather than an additional source of query complexity. It is necessary because the normalized direction is undefined at stationary points, and, when the gradient norm is arbitrarily small, the first-order term used by directional-preference comparisons can be dominated by the $L$-smoothness error. In our algorithms, $\gamma$ only determines the probing radius used in each comparison; the number of comparison queries is independent of the numerical value of $\gamma$. Once an estimate $\mathbf h\approx \nabla f(\x)/\|\nabla f(\x)\|$ is obtained, it can be used directly in normalized gradient descent updates of the form $\x_{t+1}=\x_t-\eta_t\mathbf h_t$, so the local estimation primitive can be plugged into comparison-based optimization procedures. Numerical experiment can be seen in \sec{experiments}.

For these two problems, we develop classical and quantum algorithms with comparison query complexities as follows:
\begin{itemize}[leftmargin=*]
\item Gradient testing (\algo{c-testing}): $O(1)$ queries to $O_{f}^{\mathrm{Comp}}$.
\item Classical gradient estimation (\algo{classical-estimation}): $O(n\log(1/\varepsilon))$ queries to $O_{f}^{\mathrm{Comp}}$.
\item Quantum gradient estimation (\algo{quantum-estimation}): $O(\log(n/\varepsilon))$ queries to $O_{f,Q}^{\mathrm{Comp}}$. 
\end{itemize}

In addition, we prove that our classical algorithm for gradient estimation is optimal. Another note is that our classical algorithm for gradient testing is randomized. In contrast, we prove a tight $\Theta(n)$ bound for deterministic algorithms; the corresponding algorithm and lower bound proof are detailed in \append{classical-gradient-deterministic}.

For quantum gradient estimation, we prove an $\Omega(\log(1/\varepsilon))$ lower bound, which shows that our algorithm is optimal up to a factor of $\log n$. See a summary in \tab{main}.

\begin{table*}[ht]
  \caption{Summary of our main results: classical and quantum algorithms for gradient testing and estimation by comparisons.}
  \centering
  \setlength{\tabcolsep}{8pt}
  \begin{tabular}{ccc}
     \hline
     & Gradient Testing  & Gradient Estimation \\
     \hline
     Classical & \multirow{2}{*}{$\Theta(1)$ (\thm{classical testing})} & $\Theta(n\log(1/\varepsilon))$ (\thm{classical-estimation} and \thm{classical-estimation-lower}) \\
     Quantum &  & $O(\log(n/\varepsilon))$ (\thm{quantum-estimation}), $\Omega(\log (1/\varepsilon))$ (\thm{quantum-estimation-lower}) \\
     \hline
  \end{tabular}
\label{tab:main}
\end{table*}

\section{Preliminaries}
\subsection{Notation}\label{sec:notation}

Let $f\colon\R^n\rightarrow \R $ be a differentiable $L$-smooth function, i.e.,
\begin{align}
\|\nabla f(\x)-\nabla f(\y)\|\leq L\|\x-\y\|\quad\forall\,\x,\y\in\R^{n}.
\end{align}
For any vector $\v$, $\|\v\|$ represents its $\ell_2$-norm. For any $\x\in\R^n$, we use $\g(\x)\coloneqq\nabla f(\x)/\|\nabla f(\x)\|$ to denote the normalized gradient at $\x$. We omit $\x$ when it is clear from context. We denote $\mathbf{e}_i$ to be the unit vector of the $i$-th coordinate component. We denote $S_n \coloneqq \{ \x\in\R^n \colon \|\x\| = 1\}$ to be the unit sphere in $\R^n$, and $ \x \sim S_n$ means that $\x$ is drawn from the uniform distribution on $S_n$. The orthogonal group $G(n)$ is the group of all real $n\times n$ matrices $Q$ satisfying $Q^{\top}Q = I$, which represents all orthonormal transformations of $\R^n$. The Haar measure on $G(n)$ is the unique probability measure that is invariant under left and right multiplication, and sampling from the Haar measure corresponds to choosing an orthogonal matrix uniformly at random.

\subsection{Inner Product Concentration for Random Vectors}\label{sec:inner-product-main}
Throughout the paper, we use the following estimation of the inner product between a fixed unit vector and a unit vector selected from $S_n$ uniformly at random:
\begin{lemma}\label{lem:coordinate-bound-main}
Let $n\ge 5$. For any $\x\in \mathbb{R}^n$, and any constant $c>0$, there exist constants $p_1$ and $p_2$ such that
\[
p_1 \le \Pr_{\y\sim S_n}[|\langle \y, \x\rangle| \le \|\x\|/(c\sqrt{n})]\le p_2,
\]
where $\y$ is chosen from $S_n$ uniformly at random. In particular, we have the inequalities below:
\begin{itemize}
    \item $\Pr_{\y\sim S_n}[|\langle \y, \x\rangle| \le 24\|\x\|/(25\sqrt{n})]\ge 3/5$
    \item $\Pr_{\y\sim S_n}[|\langle \y, \x\rangle| \le 18\|\x\|/(25\sqrt{n})]\le 11/20$
    \item $\Pr_{\y\sim S_n}[|\langle \y, \x\rangle| \ge \|\x\|/(5\sqrt{n})]\ge 4/5$
\end{itemize}
\end{lemma}

The lemma is adapted from Theorem 3.4.5 in \cite{vershynin2009high}. The proof of \lem{coordinate-bound-main} is deferred to \append{inner-product-append}.

\subsection{Basic Notation in Quantum Computing}\label{sec:quantum-prelim}
We use the Dirac notation $|\cdot\rangle$ to represent quantum states, which can be seen as column vectors. 
The basic unit of quantum state, qubit, is represented by a normalized vector in $\mathbb{C}^2$, which has the superposition form $|\phi\rangle = \alpha |0\rangle + \beta |1\rangle$ on the computational basis $\{|0\rangle, |1\rangle\}$ with $\alpha, \beta \in \mathbb{C}$ and $|\alpha|^2 + |\beta|^2 = 1$. An $n$-qubit system is represented in a $2^n$-dimensional complex vector space, with basis states $\{|0\rangle, |1\rangle\}^{\otimes n}$. A general $n$-qubit state can be expressed as $|\psi\rangle = \sum_{x \in \{0,1\}^n} \alpha_x |x\rangle$ with $\sum_{x} |\alpha_x|^2 = 1$. Quantum measurement on the computational basis collapses the state $|\psi\rangle = \sum_{x} \alpha_x |x\rangle$ to one of the basis states $|x\rangle$ with probability $|\alpha_x|^2$. Quantum operations are modeled as unitary transformations, which are linear operators $U$ satisfying $U^\dagger U = I$, where $U^\dagger$ is the conjugate transpose of $U$.

A quantum algorithm can access a function via queries to a quantum oracle. For a classical function $g$, the oracle $O_g$ is a unitary transformation that maps $|x\rangle|b\rangle$ to $|x\rangle|b \oplus g(x)\rangle$. This allows the oracle to be queried on superpositions of inputs, producing corresponding superpositions of outputs.
\\\\
\textbf{Quantum Fourier transform (QFT)} 
is the quantum analogue of the discrete Fourier transform, but operates on  amplitudes of quantum states. Formally, given the computational basis $\{|0\rangle, |1\rangle, \ldots, |N-1\rangle\}$, it acts as follows: 
\begin{align*}
\text{QFT} \colon |x\rangle \mapsto \frac{1}{\sqrt{N}} \sum_{k=0}^{N-1} e^{2\pi i x k / N} |k\rangle.
\end{align*}

A key application of QFT is Jordan’s gradient estimation algorithm~\cite{Jordan_2005}. Specifically, if the phase of each basis state is given by an inner product, then applying the inverse QFT and measuring in the computational basis allows one to recover the underlying vector with high probability.
\begin{proposition}[informal version of \prop{qft}]
Let $n \ge 2$ and $t \in \mathbb{Z}_+$. There exists a quantum procedure that, given an input state encoding $\x \in [0,1]^n$ as
$|\psi_\x\rangle = \frac{1}{\sqrt{(t+1)^n}} \sum_{y \in \{0,1,\dots,t\}^n} e^{2\pi i \langle \y, \x\rangle} |\y\rangle$,
outputs an estimate $\hat 
\v \in \{0, 1/t, \dots, 1\}^n$ such that
\[
\Pr\Big[\|\hat \v - \x\|_2 = O(n^{1.5}/t)\Big] \ge 2/3.
\]
Given input state $|\psi\rangle$ satisfying $\||\psi\rangle-|\psi_\x\rangle\|\le \varepsilon$, the success probability decreases by at most $O(\varepsilon)$.
\end{proposition}
Compared with Lemma 2.2 of \cite{chakrabarti2020optimization}, which gives a per-coordinate error bound for the inverse QFT, the proposition here instead bounds the overall error in $\ell_2$ norm. A formal statement with proof is deferred to \append{quantum-gradient-estimation}.

\section{Classical Algorithms}

In this section, we present our algorithms for \prob{testing} and \prob{estimation} given query access to a comparison oracle $O^{\mathrm{Comp}}_f$ defined in \eqref{eq:comparison}. We begin by developing a subroutine called ``directional preference'' that is used in both algorithms.

\subsection{Directional Preference}\label{sec:DP}
Here we show that given $\x\in \mathbb{R}^n$ and a unit vector $\v\in \mathbb{R}^n$, we can use one comparison query to understand whether the inner product $\langle\nabla f(\x), \v\rangle$ is roughly positive or negative, which determines whether $\v$ is following or against the direction of $\nabla f(\x)$. This is also known as \textit{directional preference} (DP) in~\cite{karabag2021smooth}.
\begin{lemma}\label{lem:basic oracle}
    Let $f \colon \mathbb{R}^d \to \mathbb{R}$ be an $L$-smooth function.   Given a point $\x \in \mathbb{R}^d$,   a unit vector $\v \in S_d$,   and precision $\Delta>0$ for directional preference.  Then we have:
    \begin{itemize}
        \item If $O_{f}^{\comp}(\x+\frac{2\Delta}{L}\v,  \x)=1$ ,   then $\langle\nabla  
        f(\x),  \v\rangle\geq-\Delta$.  
        \item If $O_{f}^{\comp}(\x+\frac{2\Delta}{L}\v,  \x)=-1$,   then $\langle \nabla f(\x),   \v\rangle\leq\Delta$.  
    \end{itemize}
\end{lemma}

\begin{algorithm2e}[!ht]
\caption{DP$(\x, \v, \Delta)$}
\label{algo:dp}
\LinesNumbered
\DontPrintSemicolon
\KwInput{Comparison oracle $O^{\mathrm{comp}}_f$ for $f\colon\mathbb{R}^n\to\mathbb{R}$, 
point $\x\in\mathbb{R}^n$,  unit vector $\v\in S_d$,  parameter $\Delta>0$} 
\If{$O^{\mathrm{comp}}_f\!\left(\x+\frac{2\Delta}{L}\v, \, \x\right)=1$}
    {\Return ``$\langle \nabla f(\x),  \v\rangle \ge -\Delta$"}
\Else
    {\Return ``$\langle \nabla f(\x),  \v\rangle \le \Delta$"}
\end{algorithm2e}

\begin{proof}
    Since $f$ is an $L$-smooth differentiable function,
\[
\bigl| f(\y) - f(\x) - \langle \nabla f(\x), \y - \x \rangle \bigr|
\le \frac{1}{2} L \|\y-\x\|^2
\]
for any $\x,\y \in \mathbb{R}^n$.
Take $\y = \x + \frac{2\Delta}{L}\v$, this gives
\[
f(\y) - f(\x) - \frac{2\Delta}{L}\langle \nabla f(\x), \v\rangle
\le \frac{1}{2}L\left(\frac{2\Delta}{L}\right)^2
= \frac{2\Delta^2}{L}.
\]
Therefore, if $O^{\mathrm{comp}}_f(\y,\x)=1$, i.e., $f(\y)\ge f(\x)$,
\[
\frac{2\Delta}{L}\langle \nabla f(\x), v\rangle
\ge
\frac{2\Delta}{L}\langle \nabla f(\x), \v\rangle + f(\x)-f(\y)
\ge -\frac{2\Delta^2}{L},
\]
and hence
\[
\langle \nabla f(\x), \v\rangle \ge -\Delta.
\]
On the other hand, if $O^{\mathrm{comp}}_f(\y,\x)=-1$, i.e., $f(\y)\le f(\x)$, we have
\[
\frac{2\Delta}{L}\langle \nabla f(\x), \v\rangle
\le
f(\y)-f(\x)+\frac{2\Delta^2}{L}
\le \frac{2\Delta^2}{L},
\]
and hence $\langle \nabla f(\x), \v\rangle \le \Delta$.
\end{proof}

\subsection{Gradient Testing}
Here we present our algorithm for gradient testing (\prob{testing}). Without loss of generality, we assume the given unit vector $\v$ satisfies $\v=(1,  0,  \ldots,  0)^\top$.

\begin{algorithm2e}[htbp!]
\caption{Classical Gradient Testing}
\label{algo:c-testing}
\LinesNumbered
\DontPrintSemicolon
\KwInput{testing direction $\v$, lower bound $\gamma$ on $\|\nabla f\|$} 
Set $\Delta = \varepsilon\gamma/(25\sqrt{2}n)$, $\delta = 1/3$, $T=\lceil3200\ln (1/\delta)\rceil$

Initialize $N = 0$

\For{$i=1$ \emph{to} $T$}{
Randomly choose $\y\sim S_{n-1}$ and define $\alpha_\y=(-\varepsilon/\sqrt{(n-1)(1-\varepsilon^2)}, \y)$

Call \algo{dp} with input $(\x, \alpha_\y/\|\alpha_\y\|, \Delta )$\label{lin:classical testing 1}\label{lin:call dp}

\If{$\langle\nabla f,\alpha_\y/\|\alpha_\y\|\rangle\leq\Delta$}{
$N\leftarrow N+1$
}
}

\If{$N/T\ge 63/80$}{\label{lin:yes-testing}
\Return Yes
}
\Else{\label{lin:no-testing}
\Return No
}
\end{algorithm2e}

The high-level idea of \algo{c-testing} is as follows. For our convenience, we decompose $\g$ as $\g = (g_1, \tilde{\g})$ and denote $\u = \v - \g$. Note that $\langle \tilde{\g}, \y\rangle$ is related to the length of the norm $\|\tilde{\g}\|$. If we set a comparison standard $k$, then for different $\|\tilde{\g}\|$, the probability of $\y$ such that the inner product is less than $k$ is also different. Specifically, we set $k = \varepsilon/\sqrt{n-1}$ here and in this case, there is a constant difference between the different probability. As a result, we can randomly choose a constant number of $\y$ to distinguish them.

\begin{theorem}\label{thm:classical testing}
    \algo{c-testing} solves \prob{testing} using $O(1)$ queries to a comparison oracle $O_f^{\mathrm{Comp}}$ \eqref{eq:comparison} with success probability at least $2/3$.
\end{theorem}

\begin{proof}[Proof of \thm{classical testing}]
    Since $\|\g\| = 1$, we have
    \begin{align*}
        \|\g - \e_1\|^2 = (1-g_1)^2 + g_2^2+ \dots + g_n^2=(1-g_1)^2+1-g_1^2=2-2g_1. 
    \end{align*}
    If $\|\g - \e_1\|\le \varepsilon$, we can get $g_1 \ge 1 - \varepsilon^2/2$, which means
    \begin{align}\label{eq:yes-reduce}
        \|\tilde{\g}\|^2=1 - g_1^2 \le 1 - (1-\varepsilon^2/2)^2 = \varepsilon^2 - \varepsilon^4/4 \le \varepsilon^2,
    \end{align}
    and
    \begin{align}\label{eq:yes-reduce2}
        \varepsilon\langle \v, \g\rangle/\sqrt{1-\varepsilon^2} = \varepsilon g_1/\sqrt{1-\varepsilon^2} = \varepsilon \sqrt{(1-\|\tilde{\g}\|^2)/(1-\varepsilon^2)}\ge \|\tilde{\g}\|. 
    \end{align}
    If $\|\g - \e_1\|\ge 2\varepsilon$, we can get $g_1 \le 1 - 2\varepsilon^2$, which means
    \begin{align}\label{eq:no-reduce}
         \|\tilde{\g}\|^2=\sum_{i=2}^ng_i^2 = 1 - g_1^2\ge 1 - (1 - 2\varepsilon^2)^2 = 4\varepsilon^2 - 4\varepsilon^4\geq 2\epsilon^2, 
    \end{align}
    and
\begin{align}\label{eq:no-reduce2}
    \varepsilon\langle \v, \g\rangle/\sqrt{1-\varepsilon^2}=\varepsilon \sqrt{(1-\|\tilde{\g}\|^2)/(1-\varepsilon^2)}\le \|\tilde{\g}\|/\sqrt{2}, 
\end{align}
since $\varepsilon\le1/\sqrt{2}$. Consider our task to determine whether $\|\u\|\le\varepsilon$ or $\|\u\| \ge 2\varepsilon$. By \eq{yes-reduce} and \eq{no-reduce}, it suffices to determine whether $\|\tilde{\g}\|\le\varepsilon$ or $\|\tilde{\g}\|\ge \sqrt{2}\varepsilon$. For any $\y \in S_{n-1}$, if we call \algo{dp} with input $(\x, \alpha_\y/\|\alpha_\y\|, \Delta)$ and get $\langle \nabla f(\x),  \alpha_\y/\|\alpha_\y\|\rangle \le \Delta$, we have
\begin{align*}
    \langle \y, \tilde{\g}\rangle = \langle (0, \y), \g\rangle&\le \frac{\varepsilon}{\sqrt{(n-1)(1-\varepsilon^2)}}\langle \v, \g\rangle + \Delta\|\alpha_\y\| \le \frac{\varepsilon}{\sqrt{(n-1)(1-\varepsilon^2)}}\langle \v, \g\rangle + \frac{\varepsilon}{25n}.
\end{align*}
Otherwise, we have
\[
\langle \y, \tilde{\g}\rangle \ge \frac{\varepsilon}{\sqrt{(n-1)(1-\varepsilon^2)}}\langle \v, \g\rangle - \frac{\varepsilon}{25n}.
\]
Hence we can obtain
\begin{align*}
    &\Pr_{\y\sim S_{n-1}}\left[\langle \y, \tilde{\g}\rangle \le \frac{1}{\sqrt{n-1}}\cdot\frac{\varepsilon\langle \v, \g\rangle}{\sqrt{1-\varepsilon^2}} - \frac{\varepsilon}{25n}\right]\\ 
&\qquad\le \Pr_{\y\sim S_{n-1}}\left[\text{\algo{dp} outputs ``$\langle\nabla f,\alpha_\y/\|\alpha_\y\|\rangle\leq\Delta$'' in \lin{call dp}}\right]\\
&\qquad\le \Pr_{\y\sim S_{n-1}}\left[\langle \y, \tilde{\g}\rangle
\le \frac{1}{\sqrt{n-1}}\cdot\frac{\varepsilon\langle \v, \g\rangle}{\sqrt{1-\varepsilon^2}} + \frac{\varepsilon}{25n}\right] 
\end{align*}
By \lem{coordinate-bound-main}, if $\|\u\| \le \varepsilon$, which implies $\|\tilde{\g}\|\le \varepsilon$ and \eq{yes-reduce2}, we can get
\begin{align*}
    \mathbb{E}[N/T] &\ge \Pr_{\y\sim S_{n-1}}\left[\langle \y, \tilde{\g}\rangle \le \frac{1}{\sqrt{n-1}}\cdot\frac{\varepsilon\langle \v, \g\rangle}{\sqrt{1-\varepsilon^2}} - \frac{\varepsilon}{25n}\right]\\&\ge\Pr_{\y\sim S_{n-1}}\left[\langle\y, \tilde{\g}\rangle\le \frac{\|\tilde{\g}\|}{\sqrt{n-1}} - \frac{\|\tilde{\g}\|}{25(n-1)}\right]\ge \frac{1}{2}+\frac{1}{2}\cdot\frac{3}{5}=\frac{4}{5}.
\end{align*}
By Hoeffding's inequality,
\[
\Pr[\frac{4}{5}-\frac{N}{T}\ge \frac{1}{80}]\le \mathrm{e}^{-2T/80^2}\le \delta = 1/3, 
\]
which means, from \lin{yes-testing}, we at least have the probability of $2/3$ to get the right answer Yes. Otherwise, if $\|\u\| \ge 2\varepsilon$, which implies $\|\tilde{\g}\|\ge \sqrt{2}\varepsilon$ and \eq{no-reduce2}, we have
\begin{align*}
     \mathbb{E}[N/T] &\le \Pr_{\y\sim S_{n-1}}\left[\langle \y, \tilde{\g}\rangle
\le \frac{1}{\sqrt{n-1}}\cdot\frac{\varepsilon\langle \v, \g\rangle}{\sqrt{1-\varepsilon^2}} + \frac{\varepsilon}{25n}\right]\\
&\le\Pr_{\y\sim S_{n-1}}\left[\langle\y, \tilde{\g}\rangle\le \frac{\|\tilde{\g}\|}{\sqrt{2(n-1)}} + \frac{\|\tilde{\g}\|}{25(n-1)}\right] \le \frac{1}{2}+\frac{1}{2}\cdot\frac{11}{20}=\frac{31}{40}.
\end{align*}
Similarly by Hoeffding's inequality, we have
\[
\Pr[\frac{N}{T}-\frac{31}{40}\ge \frac{1}{80}]\le \mathrm{e}^{-2T/80^2}\le \delta = 1/3, 
\]
which means, from \lin{no-testing}, we at least have the probability of $2/3$ to get the right answer No. 
\end{proof}

Finally, we note that in the deterministic setting, gradient testing has a tight query complexity of $\Theta(n)$:

\begin{theorem}\label{thm:classical-testing-lower}
The classical deterministic query complexity for solving \prob{testing} with queries to a comparison oracle $O_f^{\mathrm{Comp}}$ \eqref{eq:comparison} is $\Theta(n)$.
\end{theorem}

Specifically, given the gradient $\g$ and target direction $\v=\e_1$, our algorithm (details given in \append{upper-classical-testing}) estimates the ratio $g_i/g_1 =\langle \g, \e_i \rangle / \langle \g, \v \rangle$ for the remaining $n-1$ orthogonal directions $\e_i$ within a constant multiplicative factor. This is achieved via multiplicative search for a parameter $\beta$ such that $\g$ is nearly orthogonal to $\beta \e_1 - \e_i$, ensuring that the total query complexity across all $n-1$ directions remains $O(n)$. Then we distinguish the two instances by verifying whether $\sum_{i=2}^n |g_i/g_1|^2$ satisfies the geometric constraint of the YES case.
The $\Omega(n)$ lower bound proof (details given in \append{lower-classical-testing}) builds a hard distribution where the unknown gradient is either exactly along a target direction or is slightly ``tilted'' toward one randomly chosen orthogonal direction. Any single query almost never detects this tilt, and gradient testing has low success probability unless $\Omega(n)$ queries are made.

\subsection{Gradient Estimation}
Here we present our algorithm for gradient estimation (\prob{estimation}). As a warm-up, we first present an algorithm that finds a constant approximation of the normalized gradient.

\begin{algorithm2e}[!ht]
\caption{Classical Gradient Estimation with Constant Precision}
\label{algo:classical-first-step}
\LinesNumbered
\DontPrintSemicolon
\KwInput{lower bound $\gamma$ on $\|\nabla f\|$} 
\KwOutput{normalized vector $\u$ such that $\langle \u, \g\rangle\ge 1/10$}
Set $\Delta_1 = \gamma/n$

Let $U$ be drawn from the Haar measure on $G(n)$, where $G(n)$ denotes the orthogonal group in $\R^n$. Initialize orthonormal frame $(\v_1, \v_2,\dots, \v_n) = (U\e_1, U\e_2,\dots,U\e_n)$

\For{$i = 1$ \emph{to} $n$}{
Call \algo{dp} with input $(\x,  \v_i,  \Delta_1)$.  

\If{$\langle \v_i,  \nabla f\rangle < \Delta_1$}{\label{lin:begin-random-orthonormal}
$\v_i \leftarrow -\v_i$
}
}\label{lin:end-random-orthonormal}

\Return $\u = \frac{1}{\sqrt{n}} \sum_{i=1}^n \v_i$\label{lin:first-coordinate} 

\end{algorithm2e}

\begin{proposition}
\label{prop:classical-first-step}
    \algo{classical-first-step} outputs a normalized vector $\u$ satisfying $\langle \u, \g\rangle\ge 1/10$ using $O(n)$ queries to a comparison oracle $O_f^{\mathrm{Comp}}$ \eqref{eq:comparison} with success probability at least 2/3.  
\end{proposition}

\begin{proof}
    In \lin{end-random-orthonormal}, if $\langle \v_i, \g\rangle<\Delta_1/\gamma$, we flip $\v_i$. Hence, when the algorithm terminates we have
\begin{align}\label{eq:lower-coordinate}
\langle \v_i, \g \rangle\ge -\Delta_1/\gamma\ge -1/n.
\end{align}
Under \eq{lower-coordinate}, we can promise that the vector $\u$ in \lin{first-coordinate} satisfies $\E(\langle\u_1,  \g\rangle) \ge 0.7$ (this is formally stated as \lem{expect-constant} with full proofs deferred to \append{lower-coordinate}). Define $p = \Pr[\langle\u_1,  \g\rangle\ge1/10]$. Because of $\langle\u_1,  \g\rangle \le 1$, we can get $p\ge 2/3$ by Markov's inequality. 
\end{proof}

Building upon \algo{classical-first-step}, we obtain our \algo{classical-estimation} that solves \prob{estimation} to any given precision $\varepsilon$. 

\begin{algorithm2e}[!ht]
\caption{Classical Gradient Estimation with Arbitrary Given Precision}
\label{algo:classical-estimation}
\LinesNumbered
\DontPrintSemicolon
\KwInput{accuracy $\varepsilon$,  lower bound $\gamma$ on $\|\nabla f\|$} 

Set $\Delta_1 = \gamma/n$, $\Delta_2 = \varepsilon\gamma/400\sqrt{n}$ and define $w_i(\beta) = \beta\mathbf{e}_1 - \mathbf{e}_i$ 

Call \algo{classical-first-step} and get an estimation $\u$ satisfying $\langle \u, \g\rangle\ge1/10$

Apply an orthogonal change of basis so that $\u$ is mapped to $\e_1$. 

For any $i=2,  3,  \ldots,  n$,   initialize all $\ell_i=1$, $t_i = 1$. 

\For{$i=2$ \emph{to} $n$}{
Call \algo{dp} with input $(\x, \mathbf{e}_i, \Delta_1)$

\If{$\langle \nabla f, \mathbf{e}_i\rangle <\Delta_1$}{
$t_i \leftarrow -t_i$
}
}

Let $D = \diag(1, t_2,\dots,t_n)$ and apply the orthogonal transformation $D$ to the coordinate frame. 

\For{$i=2$ \emph{to} $n$}{

\Repeat{$\langle \nabla f, w_i(\beta)/\|w_i(\beta)\|\rangle > -\Delta_2$}{\label{lin:begin-finding-bound}
$\beta \leftarrow \ell_i/\sqrt{n}$ 

Call \algo{dp} with input $(\x, w_i(\beta)/\|w_i(\beta)\|,  \Delta_2)$

\If{$\langle \nabla f, w_i(\beta)/\|w_i(\beta)\|\rangle \le \Delta_2$}{
$\ell_i \leftarrow 2\ell_i$
}
}
\label{lin:end-finding-bound}

$\alpha_{i,1} \leftarrow -\ell_i/\sqrt{n}$, $\alpha_{i,2} \leftarrow \ell_i/\sqrt{n}$

\Repeat{$\alpha_{i,2} - \alpha_{i,1} < \varepsilon/4\sqrt{n}$}{\label{lin:begin-classical-binary-search}

$\alpha_i \leftarrow (\alpha_{i,1}+\alpha_{i,2})/2$

Call \algo{dp} with input $(\x, \kern-0.5mm w_i(\alpha_i)/\|w_i(\alpha_i)\| \kern-0.25mm ,\Delta_2)$

\If{$\langle \nabla f, w_i(\alpha_i)/\|w_i(\alpha_i)\|\rangle\le \Delta_2$}{
$\alpha_{i,1} \leftarrow \alpha_i$
}
\Else
{$\alpha_{i,2} \leftarrow \alpha_i$}

}
\label{lin:end-classical-binary-search}
}
\Return $\mathbf{h}=(\sum_i \alpha_i t_i)/\sqrt{\sum_i\alpha_i^2}$ where $\alpha_1 = 1$

\end{algorithm2e}

The intuition of \algo{classical-estimation} can be summarized as follows. To estimate the gradient direction $\g$ within error $\varepsilon$, a natural approach is to estimate each coordinate to relative accuracy $\varepsilon/\sqrt n$. In particular, we call \algo{dp} with input direction $\alpha_j \mathbf e_i - \alpha_i \mathbf e_j$ to roughly compare between $\alpha_i/\alpha_j$ and $g_i/g_j$. Assuming without loss of generality that $g_1$ has the largest magnitude among all coordinates, we can then perform a binary search over $[-1,1]$ to estimate each ratio $g_i/g_1$ to accuracy $\varepsilon/\sqrt n$.

However, the approach above requires $\Theta(n \log(n/\varepsilon))$ queries. To improve upon this, we adopt an idea similar to that used in \algo{classical-testing}, which fully exploits the upper bound on $\sum_{i=1}^n |g_i/g_1|^2$: the quantity $|g_i/g_1|$ cannot be large for all coordinates simultaneously. This is achieved by identifying an estimation $\u$ such that $\u$ and $\g$ has a constant overlap using \algo{classical-first-step}. Then, we rotate the frame so that $\u$ become the first ordinate, which ensures that the upper bound on $\sum_{i=1}^n |g_i/g_1|^2$ is not too large. Then, we compute an upper bound on each quantity $\sqrt n\cdot|g_i/g_1|$, denoted by $\ell_i$, which is obtained in Lines~\ref{lin:begin-finding-bound}–\ref{lin:end-finding-bound}. For each coordinate $i$, we perform a binary search over the interval $[-\ell_i/\sqrt n, \ell_i/\sqrt n]$ to estimate $g_i/g_1$. Since $\sum_{i=1}^n\ell_i^2/n = O(1)$, the number of iterations in our binary search $\sum_{i=1}^n \log (\ell_i/\varepsilon)$ is at most $O(n\log(1/\varepsilon))$. 

We note that the random frame does not need to be fully Haar distributed. The proof only uses the constant moment guarantee summarized in \lem{expect-constant}; hence an orthogonal $2$-design, or any sufficiently accurate approximation satisfying the same guarantee, can in principle replace the Haar-random frame.

\begin{theorem}\label{thm:classical-estimation}
    \algo{classical-estimation} solves \prob{estimation} using $O(n\log(1/\varepsilon))$ queries to a comparison oracle $O_f^{\mathrm{Comp}}$ with success probability at least $2/3$. Moreover, we can increase success probability to at least $1-\eta$ using $O(n\log(1/\varepsilon)\log(1/\eta))$ queries for any $\eta\in(0,1)$.
\end{theorem}

\begin{proof}
By \prop{classical-first-step}, in the rest of the proof, we consider the case that the event 
\begin{align}\label{eq:g1-assumption}
    g_1=\langle\u,\g\rangle\geq 1/10
\end{align}
holds. In Lines~\ref{lin:begin-finding-bound}–\ref{lin:end-finding-bound}, we call \algo{dp} with input direction $w_i(\ell_i/\sqrt{n})$, which can judge the sign of $\ell_i/\sqrt{n} - g_i/g_1$ (similar to \eq{sign-judgement}) within error $\Delta_2$. If the sign is negative, we multiply $\ell_i$ by $2$. As a result, when the loop terminates we have
    \begin{align*}
        \langle \nabla f, w_i(\ell_i&/\sqrt{n})\rangle \ge -\Delta_2, \\
        \frac{\ell_i}{\sqrt{n}}g_1 - g_i \ge -\Delta_2 &\sqrt{1+\ell_i^2/n}/\gamma \ge -\frac{10\Delta_2}\gamma,
    \end{align*}
    and similarly
    \begin{align*}
        \langle \nabla f, w_i(\ell_i&/2\sqrt{n})\rangle \le \Delta_2, \\
        \frac{\ell_i}{2\sqrt{n}}g_1 - g_i \le \Delta_2 &\sqrt{1+\ell_i^2/4n}/\gamma \le \frac{10\Delta_2}\gamma.
    \end{align*}
Using the value we set for $\Delta_2$, the error from \algo{dp} satisfies
    \[
    \frac{10\Delta_2}{\gamma \cdot g_1}\le \frac{100\Delta_2}{\gamma}\le \frac{\varepsilon}{4\sqrt{n}},
    \]
    hence we have
    \[
    |g_i/g_1|-\frac{\varepsilon}{4\sqrt{n}}\le \ell_i/\sqrt{n}\le 2|g_i/g_1|+\frac{\varepsilon}{2\sqrt{n}}. 
    \]
    As a result, in Lines~\ref{lin:begin-finding-bound}--\ref{lin:end-finding-bound}, we are promised to get the smallest $\ell_i$ which is a power of 2 and satisfies $\ell_i/\sqrt{n} > g_i/g_1$ up to a small error above. With \eq{g1-assumption} holding, for $\|\g\|=1$, we have $\sum_{i=2}^n g_i^2/g_1^2 = O(1)$. Consequently,  to find all $\ell_i$, the number of queries we need is
\[
\sum_{i=2}^n 2(1+\log\ell_i)\le\sum_{i=2}^n 2\ell_i \le \sum_{i=2}^n 2\ell_i^2 = O(n).
\]
    In Lines~\ref{lin:begin-classical-binary-search}--\ref{lin:end-classical-binary-search}, our binary search needs $O(\log (\ell_i/\varepsilon))$ iterations to reach accuracy $\varepsilon/4\sqrt{n}$. If we both have
    \[
    \langle \nabla f, w_i(\alpha_{i,1})\rangle \le \Delta_2  \qquad
    \langle \nabla f, w_i(\alpha_{i,2})\rangle \ge -\Delta_2, 
    \]
    we will then call \algo{dp} with input $\left(y, w_i((\alpha_{i,1}+\alpha_{i,2})/2\right),\Delta_2)$, which gives
    \[
    \left|\alpha_i - \frac{g_i}{g_1}\right|\le \varepsilon/(4\sqrt{n}) + \frac{\Delta_2\sqrt{1+\ell_i^2/n}}{\gamma \cdot g_1} \le \varepsilon/(2\sqrt{n}), 
    \]
    and
    \[
    \|\mathbf{h}-\g\|^2 \le 2|g_1|^2 \sum_{i=1}^n\left(\alpha_i-|g_i/g_1|\right)^2\le \varepsilon^2/2 < \varepsilon^2. 
    \]
    During binary search, we use
\[
\sum_i \log\left(\frac{4\ell_i}{\varepsilon}\right) \le n\log\frac{1}{\varepsilon} + \sum_i \log(4\ell_i)= O\left(n\log\frac{1}{\varepsilon}\right)
\]
queries. As a result, we use $O(n+n\log(1/\varepsilon)) = O(n\log(1/\varepsilon))$ queries with success probability at least $2/3$ (following that of \algo{classical-first-step}).

Finally, we show how to boost the success probability to at least $1-\eta$ with an overhead of $O(\log\frac{1}{\eta})$ in query complexity. Consider the $m=\lceil18\ln(1/\eta)\rceil$ independent base runs in the algorithm. Define $\mathbf h^{(r)}$ to be the output of the $r$-th iteration and set $\mathcal N_r=\{s\in\{1,\ldots,m\}:\|\mathbf h^{(r)}-\mathbf h^{(s)}\|\le 2\varepsilon/3\}$. We finally return any $\mathbf h^{(r)}$ with $|\mathcal N_r|\ge m/2$ and prove that it satisfies our requirements with success probability at least $1-\eta$.

Let $X_r$ be the indicator that the $r$-th candidate satisfies $\|\mathbf h^{(r)}-\g\|\le \varepsilon_0$. Since $\Pr[X_r=1]\ge2/3$, Hoeffding's inequality gives
\[
\Pr\left[\sum_{r=1}^m X_r\le m/2\right]\le \exp(-m/18)\le \eta.
\]
Conditioned on the complementary event, more than half of the candidates are good. Any two good candidates are within $2\varepsilon_0=2\varepsilon/3$ of each other, so every good candidate satisfies the selection condition $|\mathcal N_r|\ge m/2$. Thus the algorithm can select at least one candidate. Moreover, for any selected candidate $\mathbf h^{(r)}$, the set $\mathcal N_r$ intersects the set of good candidates because both sets have size greater than $m/2$ or at least $m/2$ with a strict majority of good candidates. Therefore, for some good $\mathbf h^{(s)}$, 
\[
\|\mathbf h^{(r)}-\g\|\le \|\mathbf h^{(r)}-\mathbf h^{(s)}\|+\|\mathbf h^{(s)}-\g\|\le 2\varepsilon/3+\varepsilon/3=\varepsilon.
\]
The total query complexity is $m\cdot O(n\log\frac{1}{\varepsilon})=O(n\log\frac{1}{\varepsilon}\log\frac{1}{\eta})$.
\end{proof}

\section{Quantum Algorithm for Gradient Estimation}

In this section, we present our quantum algorithm for gradient estimation.

\begin{algorithm2e}
\caption{Quantum Gradient Estimation}
\label{algo:quantum-estimation}
\LinesNumbered
\DontPrintSemicolon
\KwInput{$\x\in\mathbb{R}^d$,  accuracy $\varepsilon$,  lower bound $\gamma$ on $\|\nabla f\|$} 

Set $\Delta = \gamma\varepsilon^2/(48\pi n^{3})$ and the quantum transformation unitary $U$ such that $U|\y\rangle = \frac{1}{\sqrt{N}}\sum_{\z \in M} \e^{-2\pi i \langle \z,\y\rangle t}|\z\rangle$

Randomly sample a unit vector $\v$,  and rotate the frame so that $\v$ becomes the first basis vector $\mathbf{e}_1$

Initialize $|\psi\rangle=\frac{1}{\sqrt{N}}\sum_{\y\in M}
|\y\rangle|\phi\rangle$,   where $M=\{\y\mid y_i = 0, \frac{1}{t}, \frac{2}{t}, \dots, 1\}$, $t=\lceil10n^2/\varepsilon\rceil$, and $|\phi\rangle = \frac{1}{\sqrt{t^2+1}}\sum_{j=0}^{t^2} \e^{-2\pi ij/t}|j\rangle$

Set $k_1 = -5n$, and $k_2 = 5n$

\Repeat{$k_2-k_1<\varepsilon^2/(8\pi n^{1.5})$}{\label{lin:begin-quantum-binary-search}
Set $k = (k_1+k_2)/2$

For each $\y$, call \algo{dp} with input $(\x,  (k,  \tilde{\y})/\|(k, \tilde{\y})\|,  \Delta)$

\If{$\langle \nabla f, (k,  \tilde{\y})/\|(k, \tilde{\y})\|\rangle < \Delta$}{
Set $k_1 = k$
}
\Else{
Set $k_2 = k$
}
}
\label{lin:end-quantum-binary-search}

Set $h(\y) = y_1 - k$, and output $\lfloor h(\y)\cdot t^2/\sqrt{n}\rfloor$ on the second register\label{lin:phase-kickback}

Apply the quantum transformation $U$ on our first register.\label{lin:quantum-transformation}\kern-2mm

\Return the measurement result of the first register in the computational basis.
\end{algorithm2e}

As introduced in \sec{quantum-prelim}, Ref.~\cite{Jordan_2005} assumed quantum query access to $\langle \y, \nabla f\rangle$ for each $\y$ and then use QFT to estimate $\nabla f$. However, we cannot directly estimate $\langle \y, \g\rangle$ in our comparison setting. Instead, we choose a direction $\mathbf{e}_1$ and approximate $\langle \y, \g\rangle/\langle \mathbf{e}_1, \g\rangle$. To achieve this, we apply \algo{dp} with input direction $\v = (k, \tilde{\y})$, 
where we assume $\y = (y_1,\tilde{\y})$ and apply binary search on $k$ such that $\v$ is roughly orthogonal to $\g$.  
This indicates that the overlap between $\tilde{\y}$ and $\g$ is about $-k\langle \g, \mathbf{e}_1\rangle$.
Then, we add $y_1$, obtain $(y_1 - k)\langle \mathbf{e}_1, \g\rangle$ as $\langle \y, \g\rangle$, and kick this back to the phase. Finally, we apply QFT to get the estimation.

\begin{theorem}\label{thm:quantum-estimation}
\algo{quantum-estimation} solves \prob{estimation} using $O(\log(n/\varepsilon))$ queries to a quantum comparison oracle $O_{f,Q}^{\mathrm{Comp}}$ \eqref{eq:comparison-oracle-quantum} with success probability at least $8/15-2\varepsilon$. Moreover, we can increase success probability to $1-\eta$ using $O(\log\frac{n}{\varepsilon}\log\frac{1}{\eta})$ queries for any $\eta\in(0,1)$. 
\end{theorem}

\begin{proof}
    By \lem{coordinate-bound-main}, we have 
    \[
    \Pr_{\v\sim S_n}[|\langle \v, \g\rangle|\ge1/(5\sqrt{n})] \ge 4/5.
    \]
    Therefore, without loss of generality, we can assume $|\langle \v, \g\rangle|\ge1/(5\sqrt{n})$, which means for any $\y\in M$, 
    \[
    |\langle \g, \y\rangle|\le \sqrt{n} \le 5n|\langle \g, \v\rangle|, 
    \]
    indicating that there exists $k\in[-n, n]$ such that
    \[
    \langle \g, (k, \tilde{\y})\rangle = 0.
    \]
    During our binary search in Lines~\ref{lin:begin-quantum-binary-search}--\ref{lin:end-quantum-binary-search}, for each $\y$, define $\alpha_\y(k) = (k, \tilde{\y})$. If we already obtain
    \begin{align*}
    \langle \nabla f, \alpha_\y(k_1)/\|\alpha_\y(k_1)\|\rangle&\le \Delta \\
    \langle \nabla f, \alpha_\y(k_2)/\|\alpha_\y(k_2)\|\rangle&\ge -\Delta, 
    \end{align*}
    we will then call \algo{dp} with input $(\x, \alpha_\y((k_1+k_2)/2)/\|\alpha_\y((k_1+k_2)/2)\|,  \Delta)$. Finally, after $O(\log (n/\varepsilon))$ iterations, we get
    \begin{align*}
        \langle \nabla f, (k , \tilde{\y})\rangle \le \|(k , \tilde{\y})\|\Delta &\le(5n+n)\Delta = 6n\Delta, \\
        \langle \nabla f, (k + \varepsilon^2/(8\pi &n^{1.5}), \tilde{\y})\rangle \ge -6n\Delta, 
    \end{align*}
    which also means
    \begin{align*}
         \langle \g, \y \rangle &\le (y_1 - k)\langle \g, \v \rangle + 6n\Delta/\gamma, \\
         \langle \g, \y \rangle \ge (y_1 &- k - \varepsilon^2/(8\pi n^{1.5}))\langle \g, \v \rangle - 6n\Delta/\gamma. 
    \end{align*}
    And then we have
    \[
    \left|h(\y) - \frac{\langle \g, \y\rangle}{\langle \g, \v\rangle}\right|\le \varepsilon^2/(8\pi n^{1.5}) + \frac{6n\Delta}{\gamma\langle\g, \v\rangle} \le \varepsilon^2/(4\pi n^{1.5}), 
    \]
    which means
    \[
    \left|\e^{2\pi i h(\y) \cdot \frac{t}{5\sqrt{n}}} - \e^{2\pi i \cdot \langle\frac{ \g }{\langle \g, \v\rangle}, \y\rangle \cdot \frac{t}{5\sqrt{n}}}\right| \le |\e^{i\varepsilon} - 1| \le \varepsilon. 
    \]
    In \lin{phase-kickback}, we can get the quantum state
    \[
    |\psi\rangle = \frac{1}{\sqrt{N}}\sum_{\y\in M}\e^{2\pi i h(\y)\cdot \frac{t}{5\sqrt{n}}}|\y\rangle|\phi\rangle.
    \]
    As a result,  we get
    \[
    \bigg\||\psi\rangle-\frac{1}{\sqrt{N}}\sum_{\y\in M} \e^{2\pi i \langle\frac{ \g }{\langle \g, \v\rangle}, \y\rangle\cdot \frac{t}{5\sqrt{n}}}|\y\rangle\bigg\|\le \varepsilon.  
    \]
    By \prop{qft},  we can output the result $\z$ after the transformation in \lin{quantum-transformation} such that
    \[
    \left\|\z - \frac{ \g }{5\sqrt{n}\cdot\langle \g, \v\rangle}\right\| \le 2n^{1.5}/t \le\varepsilon/(10\sqrt{n})
    \]
with probability at least $2/3 - 2\varepsilon$. This further implies
\[
\left\|\frac{\z}{\|\z\|} - \g\right\|\le 2\sin\theta\le10\sqrt{n}\left\|\z - \frac{ \g }{5\sqrt{n}\cdot\langle \g, \v\rangle}\right\|\le \varepsilon, 
\]
where $\theta \!= \!\arg(\frac{\z}{\|\z\|}, \g)$. In all, we can succeed with probability at least $\frac{4}{5}\cdot(\frac{2}{3}-2\varepsilon)\ge\frac{8}{15} - 2\varepsilon$. 

In addition, the success probability can also be boosted to $1-\eta$ with any $\eta \in (0,1)$ similar to the proof of \thm{classical-estimation}. 
\end{proof}

\section{Numerical Experiments}\label{sec:experiments}
We complement our theoretical results with numerical experiments that (i)~validate the empirical query complexity of the proposed algorithms, and (ii)~demonstrate that plugging our gradient direction estimator into a NGD optimizer yields convergence comparable to access to the exact normalized gradient, and better than existing comparison based baselines under matched per-iteration query budgets. All experiments are run on an Apple MacBook Air (M2, 2022) with 8-core Apple M2 chip and 16 GB memory.

\paragraph{Setup.}
We use three test functions: a general strongly-convex quadratic $f(\x)=\tfrac12\x^{\!\top} U^{\!\top}\Lambda U\x$ with $U$ Haar-orthogonal and $\Lambda$ a diagonal matrix whose eigenvalues are evenly spread over $[1,10]$; a sparse quadratic function in which the first $s=10$ coordinates carry weights evenly spread over $[1,10]$ and the remaining coordinates carry a vanishing weight; and the extended Rosenbrock function $f(\x)=\sum_{i=1}^{n-1}\big[100(x_{i+1}-x_i^2)^2+(1-x_i)^2\big]$. 

\paragraph{Correctness of our algorithms.}
We evaluate the three classical algorithms developed in this paper: the randomized $O(1)$ gradient testing algorithm (\algo{c-testing}), the deterministic $O(n)$ gradient testing algorithm (\algo{classical-testing}), and the gradient estimation algorithm (\algo{classical-estimation}).
For testing, we construct an extreme YES instance at distance $0.95\,\varepsilon$ from $\nabla f(\x)/\|\nabla f(\x)\|$ and an extreme NO instance at distance $2.05\,\varepsilon$, and run $100$ independent trials for each $(n,\varepsilon)$ configuration with $n\in\{10,50,100\}$ and $\varepsilon=0.2$. Both testing algorithms attain an empirical success probability of $100\%$ on every configuration, far exceeding the constant probability guarantee in our theoretical analysis.
For estimation, \tab{est-dim} reports the dependence on dimension at $\varepsilon=0.2$ across all three test functions, and \tab{est-eps} reports the dependence on the target precision $\varepsilon$ on the extended Rosenbrock function at $n=100$. The estimator returns a unit vector whose error is below $\varepsilon$ in nearly $100\%$ of trials, with average error well below the target precision. This confirms that the worst-case query bound is not tight in typical instances.
\begin{table}[t]
  \centering
  \small
  \caption{Estimation accuracy and average query count for \algo{classical-estimation} across dimensions, $\varepsilon=0.2$, $100$ trials per cell. Success $=$ fraction of trials with $\ell_2$ error below $\varepsilon$.}
  \label{tab:est-dim}
  \begin{tabular}{llccc}
    \toprule
    Function & $n$ & Success & Avg.\ error & Avg.\ queries \\
    \midrule
    \multirow{3}{*}{Gen.\ Quadratic}
      & 10  & 1.00 & 0.0068 & 85.3  \\
      & 50  & 1.00 & 0.0071 & 461.0 \\
      & 100 & 1.00 & 0.0071 & 929.6 \\
    \midrule
    \multirow{3}{*}{Sparse Quadratic}
      & 10  & 1.00 & 0.0068 & 86.4  \\
      & 50  & 1.00 & 0.0071 & 458.5 \\
      & 100 & 1.00 & 0.0071 & 922.3 \\
    \midrule
    \multirow{3}{*}{Ext.\ Rosenbrock}
      & 10  & 1.00 & 0.0068 & 85.4  \\
      & 50  & 1.00 & 0.0072 & 459.8 \\
      & 100 & 1.00 & 0.0072 & 925.7 \\
    \bottomrule
  \end{tabular}
\end{table}
\begin{table}[t]
  \centering
  \small
  \caption{Estimation accuracy and average query count for \algo{classical-estimation} on the extended Rosenbrock function at $n=100$ as the target precision $\varepsilon$ varies, $100$ trials per cell.}
  \label{tab:est-eps}
  \begin{tabular}{ccccc}
    \toprule
    $\varepsilon$ & Success & Avg.\ error & Max.\ error & Avg.\ queries \\
    \midrule
    0.20 & 1.00 & 0.0072  & 0.0081  & 925.7  \\
    0.10 & 1.00 & 0.0036  & 0.0041  & 1024.9 \\
    0.05 & 1.00 & 0.0018  & 0.0020  & 1124.5 \\
    0.01 & 1.00 & 0.00045 & 0.00052 & 1320.7 \\
    \bottomrule
  \end{tabular}
\end{table}

\begin{figure*}[t]
  \centering
  \begin{subfigure}{0.32\linewidth}
    \includegraphics[width=\linewidth]{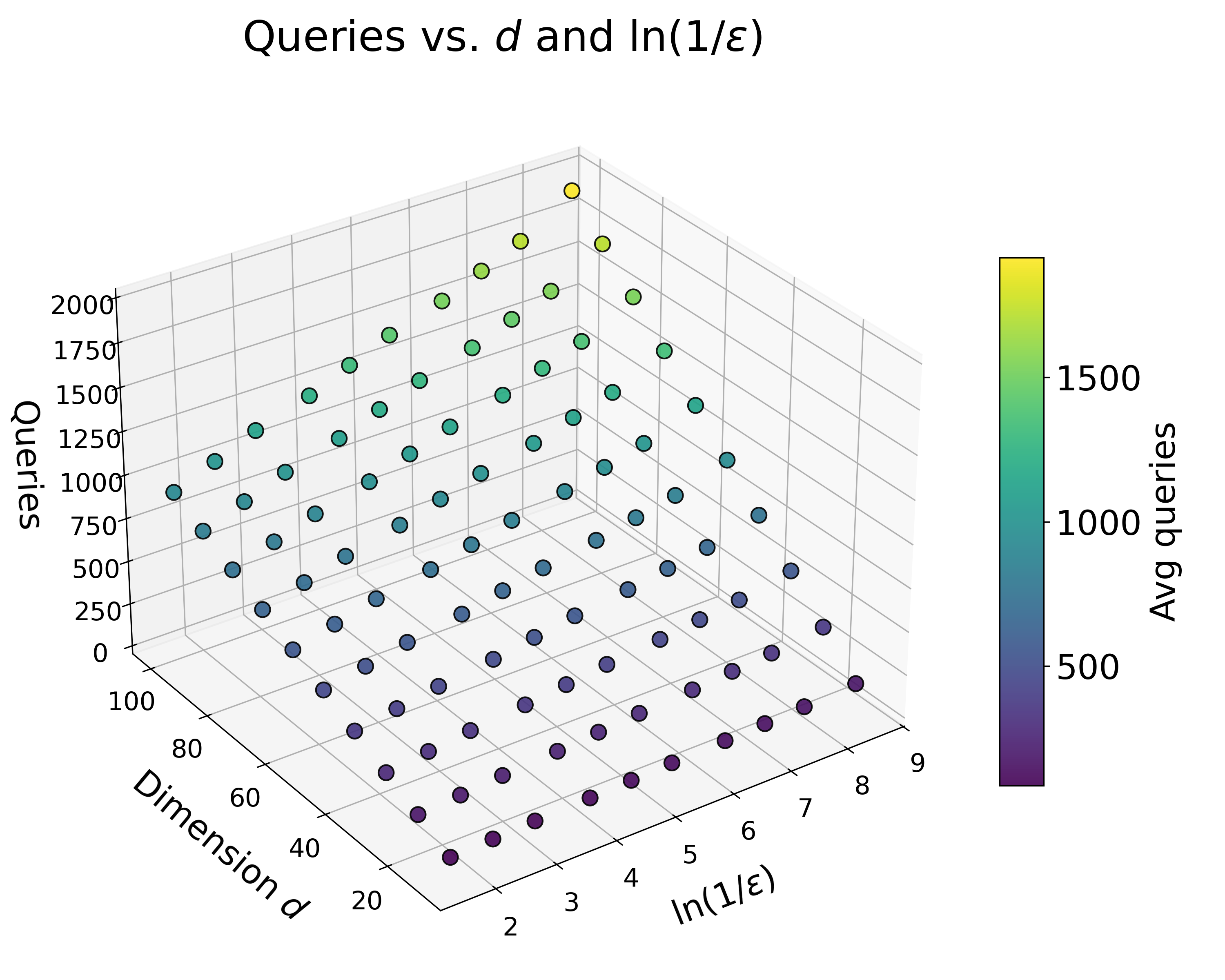}
    \caption{Empirical query complexity.}
    \label{fig:exp-complexity}
  \end{subfigure}\hfill
  \begin{subfigure}{0.32\linewidth}
    \includegraphics[width=\linewidth]{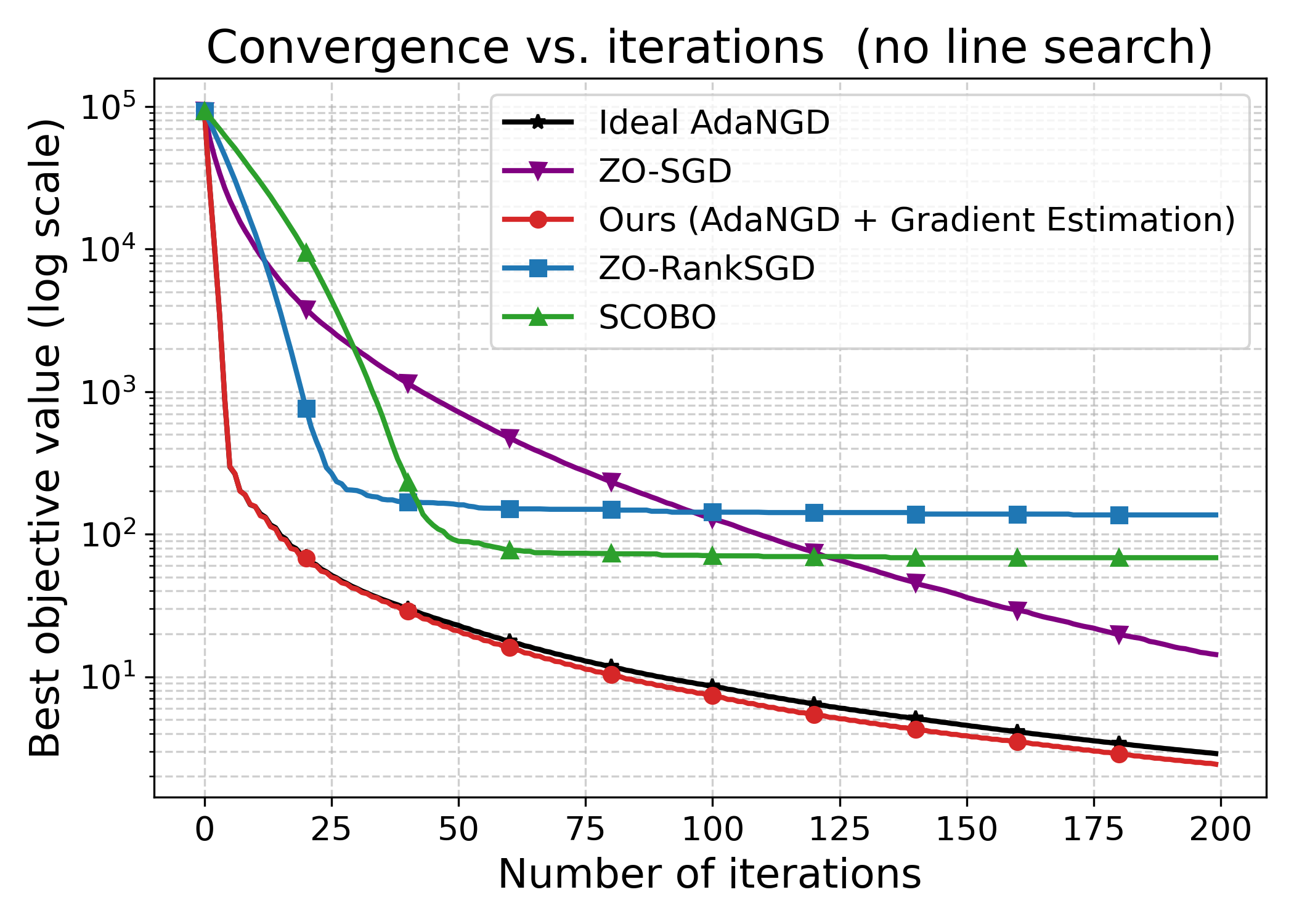}
    \caption{Convergence, no line search.}
    \label{fig:exp-no-ls}
  \end{subfigure}\hfill
  \begin{subfigure}{0.32\linewidth}
    \includegraphics[width=\linewidth]{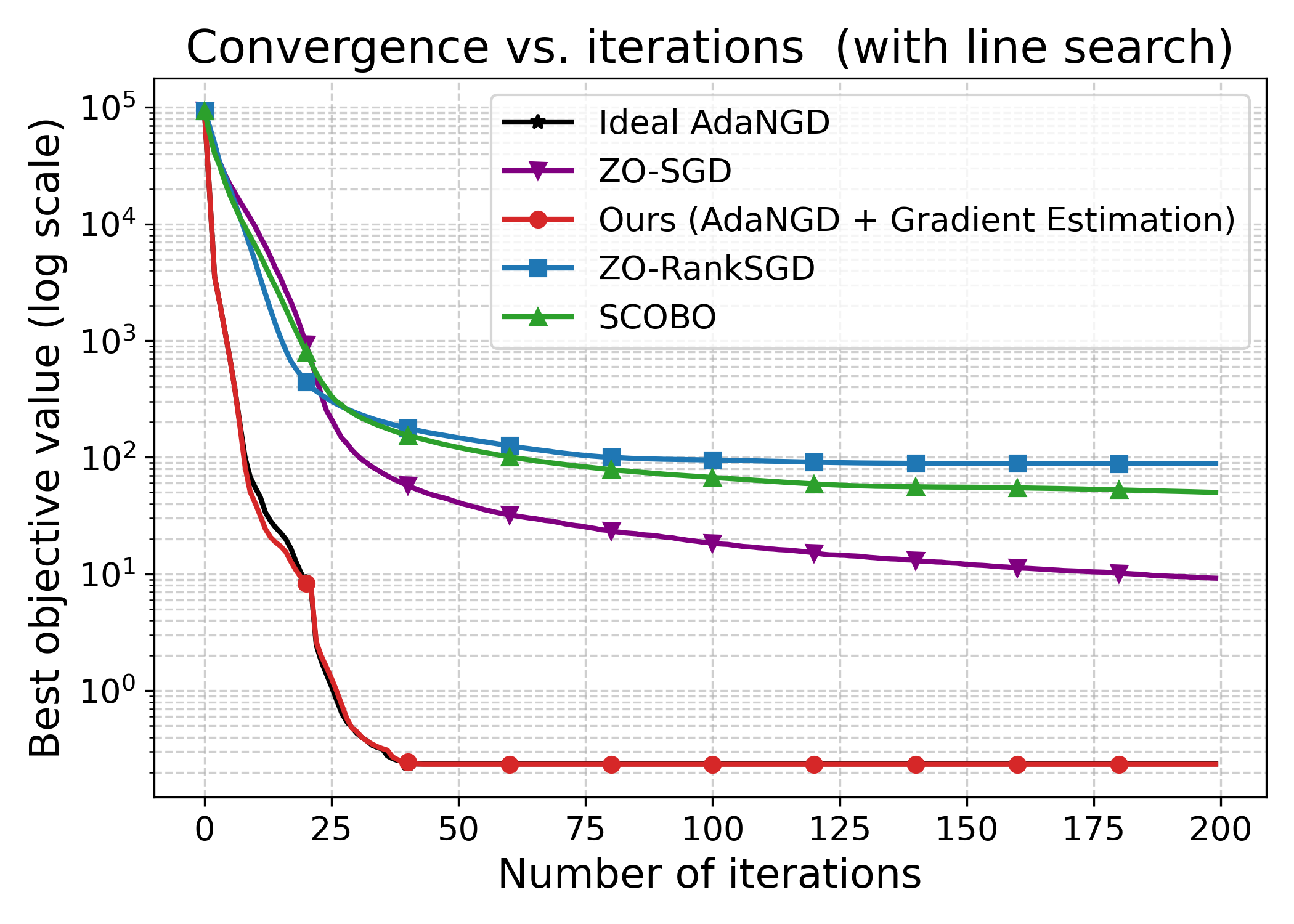}
    \caption{Convergence, with line search.}
    \label{fig:exp-with-ls}
  \end{subfigure}
  \caption{(a)~Average comparison queries used by \algo{classical-estimation} on the extended Rosenbrock function over a $10\times 10$ grid of $(n,\ln(1/\varepsilon))$ values; the fitted bilinear surface is $Q\approx 1.38\,n\ln(1/\varepsilon)+7.06\,n-1.41\ln(1/\varepsilon)-5.28$ with $R^2=0.999$. (b),~(c)~Convergence on the extended Rosenbrock function in $n=100$, averaged over $10$ random initializations, without and with line search respectively. Under per-iteration comparison budgets of the same order, \textsc{AdaNGD$+$Ours} tracks the noise-free \textsc{Ideal AdaNGD} curve, while the baselines stall orders of magnitude above the optimum.}
  \label{fig:experiments}
\end{figure*}

\paragraph{Empirical query complexity.}
\thm{classical-estimation} predicts $Q=O(n\log(1/\varepsilon))$ comparison queries for our estimation algorithm. We verify this by sweeping a $10\times 10$ grid of $(n,\varepsilon)$ values on the extended Rosenbrock function with $\gamma=0.05$ and $100$ trials per cell. \fig{exp-complexity} shows the joint dependence of the average query count on $n$ and $\ln(1/\varepsilon)$, following the bilinear model
\[
\textstyle Q \approx w_1\,n\ln(1/\varepsilon)+w_2\,n+w_3\ln(1/\varepsilon)+b.
\]
Fitting this model jointly across the grid gives $w_1\approx 1.38$, $w_2\approx 7.06$, $w_3\approx -1.41$, $b\approx -5.28$ with $R^2=0.999$, in tight agreement with the theoretical scaling.

\paragraph{Optimization with the estimated direction.}

We feed the direction returned by \algo{classical-estimation} into adaptive NGD \citep{levy2017online} with step size $\eta_t=R/\sqrt{2t}$, $R=5$, giving \textsc{AdaNGD$+$Ours}. Baselines on the extended Rosenbrock function in $n=100$ (averaged over $10$ random initializations) are:
\textsc{Ideal AdaNGD}, the noise-free reference using the true normalized gradient;
\textsc{ZO-SGD}~\citep{nesterov2017random}, a two-point Gaussian smoothing estimator with $m=100$ probes;
\textsc{ZO-RankSGD}~\citep{tang2023zeroth}, an $(m,k)$ ranking oracle with $m=k=100$;
\textsc{SCOBO}~\citep{cai2022one}, $m=100$ pairwise sign comparisons per iteration.
We take $\varepsilon=0.2$ for our estimator, where the cost is of the same order as the baselines. We also report a line-search variant ($\ell=5$ shrinking trials, factor $0.1$). The results are shown in~\fig{exp-no-ls} and~\fig{exp-with-ls}.

After $200$ iterations \textsc{AdaNGD$+$Ours} reaches objective $0.23$ with line search and $2.6$ without, essentially matching \textsc{Ideal AdaNGD} ($0.23$ and $3.1$); while the other three baselines still remain over $10$. At a comparable per-iteration query budget, our estimator returns a direction whose angular deviation is small enough that adaptive NGD behaves as if it had exact gradient direction, whereas the other baselines spend the same budget on a much coarser direction.

\section{Lower Bounds on Gradient Estimation}
We also establish lower bounds on gradient estimation, both in the classical and quantum settings.

\begin{theorem}\label{thm:classical-estimation-lower}
Any classical algorithm that solves \prob{estimation} with success probability $2/3$ takes $\Omega(n\log(1/\varepsilon))$ queries to a comparison oracle $O_f^{\mathrm{Comp}}$ \eqref{eq:comparison}.
\end{theorem}

The proof is built upon an $\varepsilon$-net argument on the sphere, and a gradient estimator must effectively identify which one is present. Each comparison query yields only a constant number of outcomes, conveys limited information, implying that the number of queries should scale linearly with dimension. The detailed proof can be found in \append{lower-classical-estimation}.

\begin{theorem}\label{thm:quantum-estimation-lower}
Any quantum algorithm that solves \prob{estimation} with success probability $2/3$ takes $\Omega(\log (1/\varepsilon))$ queries to a quantum comparison oracle $O_{f,Q}^{\mathrm{Comp}}$ \eqref{eq:comparison-oracle-quantum}.
\end{theorem}

This quantum lower bound argument applies the hybrid argument~\cite{bennett1997strengths}, where we consider a hard instance with binary partial derivatives and prove that we need enough number of quantum queries to estimate the gradient. The detailed proof can be found in \append{lower-quantum-estimation}.

\section{Conclusions}
In this paper, we study gradient testing and gradient estimation using comparison queries. For any smooth $f\colon\mathbb R^n\to\mathbb R$, $\x\in\R^n$, and $\varepsilon>0$, our classical gradient testing algorithm determines whether $\nabla f(\x)/\|\nabla f(\x)\|$ is $\varepsilon$-close or $2\varepsilon$-far from a given unit vector $\v$ using $O(n)$ queries, as well as a gradient estimation algorithm that outputs an $\varepsilon$-estimate of $\nabla f(\x)/\|\nabla f(\x)\|$ using $O(n\log(1/\varepsilon))$ queries. Both bounds are shown to be optimal. Furthermore, we study these problems in the quantum setting, and develop quantum algorithms for gradient testing and gradient estimation using $O(1)$ and $O(\log (n/\varepsilon))$ queries, respectively. 

Technically, after an appropriate rotation, there exists a dominant coordinate whose inner product with the gradient captures a constant fraction of the gradient magnitude. We isolate this coordinate and express inner product as a scale of this constant fraction, and also optimized the application of binary search in other coordinates to achieve optimality. In terms of quantum algorithms, we leverage the quantum Fourier transform (QFT) to coherently encode directional derivatives across coordinates into relative phases of a single quantum state. This phase encoding enables the simultaneous extraction of relative inner products via inverse QFT. 

Our work raises several open directions for future research. First, our quantum algorithm for gradient estimation is not yet optimal, and we conjecture that there exists an algorithm matching our lower bound. Second, it is also natural to study the gradient testing and estimation problems with a stochastic comparison oracle. Third, it would be valuable to investigate how the proposed gradient testing and estimation procedures can be incorporated into optimization solvers with improved performance.

\section*{Acknowledgments}
We thank Xingyu Zhao for helpful discussions at the early stage of this paper. We thank the anonymous reviewers for their constructive feedback. X.T., H.W., Y.Z., and T.L. were supported by the National Natural Science Foundation of China (Grant Numbers 62372006 and 92365117).

\newcommand{\arxiv}[1]{arXiv:\href{https://arxiv.org/abs/#1}{\ttfamily{#1}}\?}\newcommand{\arXiv}[1]{arXiv:\href{https://arxiv.org/abs/#1}{\ttfamily{#1}}\?}\def\?#1{\if.#1{}\else#1\fi}

\newpage
\appendix
\onecolumn
\section{Auxiliary Lemmas for Classical Upper Bounds}

\subsection{Inner Product Concentration for Random Vectors on Sphere}\label{append:inner-product-append}
In this subsection, we give a formal proof of the inner product concentration for random vectors on sphere stated in \sec{inner-product-main}, restated below:
\begin{lemma}\label{lem:coordinate-bounds}
Let $n\ge 5$. For any $\x\in \mathbb{R}^n$, $\x \ne \mathbf{0}$, and any constant $c>0$, there exists constant $p_1$ and $p_2$ which is independent of $n$, such that
\[
p_1 \le \Pr_{\y\sim S_n}[|\langle \y, \x\rangle| \le \|\x\|/(c\sqrt{n})]\le p_2.
\]
where $\y$ is chosen from $S_n$ uniformly at random. In particular, we have the inequalities below:
\begin{itemize}
    \item $\Pr_{\y\sim S_n}[|\langle \y, \x\rangle| \le \|\x\|\cdot\frac{24}{25\sqrt{n}}]\ge 3/5$
    \item $\Pr_{\y\sim S_n}[|\langle \y, \x\rangle| \le \|\x\|\cdot\frac{18}{25\sqrt{n}}]\le 11/20$
    \item $\Pr_{\y\sim S_n}[|\langle \y, \x\rangle| \ge \|\x\|/(5\sqrt{n})]\ge 4/5$
\end{itemize}
\end{lemma}

\begin{proof}
The existence of an absolute constant $p_1>0$ such that
\[
\Pr_{\y \sim \mathrm{Unif}(S_n)}\left[ |\langle \y,\x\rangle| \le \frac{\|\x\|}{c\sqrt{n}} \right] \ge p_1
\]
for all $n \ge 5$ and all $\x \ne 0$ follows from standard anti-concentration results for one-dimensional marginals of the uniform distribution on the sphere; see Theorem 3.4.5 in \cite{vershynin2009high}. We focus on establishing the remaining bounds and the stated explicit constants.

By rotational invariance of the uniform distribution on $S_n$, there exists an orthogonal matrix $Q$ such that $Q\x = \|\x\| \e_1$. Since $Q\y \sim \mathrm{Unif}(S_n)$ and $\langle \y, \x\rangle = \langle Q\y, Q\x\rangle$, we assume without loss of generality that $\x = \|\x\| \e_1$. Then
\[
\frac{\langle \y,\x\rangle}{\|\x\|} = y_1.
\]
Define the rescaled coordinate $T_n := \sqrt{n}\, y_1$. The marginal density of $y_1$ is
\[
f_n(t) = c_n (1 - t^2)^{\frac{n-3}{2}}, \quad t \in [-1,1], \qquad 
c_n = \frac{\Gamma(n/2)}{\sqrt{\pi}\,\Gamma((n-1)/2)}.
\]
By the change of variables $t = s/\sqrt{n}$, the density of $T_n$ is
\[
h_n(s) = c_n \sqrt{n} \left(1 - \frac{s^2}{n}\right)^{\frac{n-3}{2}} \mathbf{1}\{|s| \le \sqrt{n}\}.
\]
For $t \ge 0$, define $F_n(t) := \Pr(|T_n| \le t) = 2 \int_0^t h_n(s)\d s$. For each fixed $s \in [0,1]$, $h_n(s)$ is nondecreasing in $n$ for all $n \ge 5$, which implies that $F_n(t)$ is nondecreasing in $n$ for every fixed $t \in [0,1]$. On the other hand, letting $g \sim \mathcal{N}(0, I_n)$ and $\y = \g/\|\g\|$, we have $\y \sim \mathrm{Unif}(S_n)$ and
\[
T_n = \frac{g_1}{\sqrt{\frac{1}{n}\sum_{i=1}^n g_i^2}}.
\]
By the law of large numbers and Slutsky's theorem, $T_n$ converges in distribution to $Z \sim \mathcal{N}(0,1)$, hence $F_n(t) \to 2\Phi(t)-1$ for every fixed $t \ge 0$. Using monotonicity and the explicit density for $n=5$, one obtains
\[
F_n(0.96) \ge F_5(0.96) \ge \frac{3}{5},
\]
which proves the first displayed inequality. Similarly,
\[
F_n(0.72) \le \lim_{m\to\infty} F_m(0.72) = 2\Phi(0.72) - 1 < \frac{11}{20},
\]
and
\[
\Pr(|T_n| \ge 1/5) = 1 - F_n(1/5) \ge \lim_{m\to\infty} 2(1-\Phi(1/5)) > \frac{4}{5},
\]
which completes the proof.
\end{proof}

\subsection{Constant Overlap via Basis Averaging}\label{append:lower-coordinate}
In this subsection, we prove that for a fixed unit vector in $\R^n$, assuming that its inner product with all basis vectors of a Haar-random orthonormal basis is at least $-1/n$, the given vector has a constant overlap with the uniform average of all basis vectors. Similar asymptotic bounds are standard in high-dimensional spherical measures; see, e.g., \cite{chatterjee2017uniform,kim2018conditional}.
\begin{lemma}
\label{lem:expect-constant}
Fix $n\ge 3$ and a unit vector $\v\in\mathbb R^n$.
Let $(\u_1,\dots,\u_n)$ be a Haar-random orthonormal basis in $\mathbb R^n$,
conditioned on the event
\[
\textsf{E}_n:=\Big\{\langle \u_i,\v\rangle \ge -\frac1n,\ \forall i\in[n]\Big\}.
\]
Define
\[
W_n:=\frac1{\sqrt n}\left\langle\sum_{i=1}^n \u_i,\v\right\rangle.
\]
Then for all $n\ge 500$,
\[
\mathbb E[W_n\mid \textsf{E}_n]\ >\ 0.7.
\]
\end{lemma}

\begin{proof}
Denote
\[
X:=U^\top \v,\qquad X_i=\langle \u_i,\v\rangle .
\]
Since $U$ is Haar-distributed and $\v$ is fixed, $X$ is uniformly distributed on
$S^{n-1}$. Thus
\[
W_n=\frac1{\sqrt n}\sum_{i=1}^n X_i,
\qquad
\mathsf E_n=\left\{X_i\ge -\frac1n\ \forall i\in[n]\right\}.
\]
Now let $G=(G_1,\dots,G_n)\sim N(0,I_n)$. From Eq.~3.15 of \cite{vershynin2009high}, $G/\|G\|$ is also uniformly distributed on $S^{n-1}$. Therefore,
\[
\mathbb E[W_n\mid \mathsf E_n]
=
\mathbb E\left[
\frac{\sum_{i=1}^n G_i}{nR}
\ \middle|\ 
G_i\ge -\frac{R}{\sqrt n}\ \forall i
\right].
\]
Define $\alpha_n:=1/\sqrt n$. Let
\[
m(\alpha):=\frac{\varphi(\alpha)}{\Phi(\alpha)},
\qquad
\varphi(t)=\frac{1}{\sqrt{2\pi}}e^{-t^2/2},\qquad
\Phi(t)=\int_{-\infty}^t\varphi(s)\,ds.
\]
A standard Taylor expansion at $\alpha=0$ gives
\[
m(\alpha)\ge \sqrt{\frac{2}{\pi}}-\frac{2}{\pi}\alpha
\qquad\text{for all } \alpha\in[0,1].
\]
In particular, for all $n\ge 3$,
\[
m(\alpha_n)\ge \sqrt{\frac{2}{\pi}}-\frac{2}{\pi}\frac{1}{\sqrt n}.
\]
Next, to control the error introduced by replacing $R^{-1}$ with $1$, recall that 
\[
R=\frac{\|G\|}{\sqrt n},\qquad W_n=\frac{\frac1n\sum_{i=1}^n G_i}{R}, 
\]
and we can get
\[
|W_n-m(\alpha_n)|=|\frac{\sum_{i=1}^n G_i}{nR}-\frac{\sum_{i=1}^n G_i}{n}|=\frac{\sum_{i=1}^n G_i}{n}|R^{-1}-1|\le\frac{\sqrt{\sum_{i=1}^n G_i^2}}{\sqrt{n}}|R^{-1}-1|=|R-1|.
\]
So we only need to bound $\mathbb E\left[|R-1|\mid \textsf{E}_n\right]$ explicitly. Note that $R$ is independent with $\textsf{E}_n$, we have
\[
\mathbb E\left[|R-1|\mid \textsf{E}_n\right]=\mathbb E\left[|R-1|\right]. 
\]
Since $R=\|G\|/\sqrt{n}$ with $\|G\|^2\sim\chi^2_n$, where $\chi^2_n$ is the $n$-dimension chi-square distribution, we have
$\mathrm{Var}(R)\le 1/n$ for all $n\ge 3$, and thus by Cauchy-Schwarz inequality we have
\[
\mathbb E[|R-1|]\le \sqrt{\mathrm{Var}(R)}\le \frac{1}{\sqrt {2n}}.
\]
Consequently,
\[
\mathbb E[W_n\mid \textsf{E}_n]\ \ge\ m(\alpha_n)-\frac{1}{\sqrt{2n}}
\qquad\text{for all } n\ge 3.
\]
Combining the two inequalities,
\[
\mathbb E[W_n\mid \textsf{E}_n]
\ge
\sqrt{\frac{2}{\pi}}
-\left(\frac{2}{\pi}+\frac{1}{\sqrt{2}}\right)\frac{1}{\sqrt n}
\ge
\sqrt{\frac{2}{\pi}}-\frac{1.35}{\sqrt n},
\]
Finally, if $n\ge 225$ then $1.35/\sqrt n\le 0.09$, which gives $\mathbb E[W_n\mid \textsf{E}_n]\ge0.7$.
\end{proof}

\section{Classical Deterministic Gradient Testing}
\label{append:classical-gradient-deterministic}
In this section, prove the tight $\Theta(n)$ bound for classical deterministic gradient testing.

\subsection{Classical Deterministic Gradient Testing Algorithm}\label{append:upper-classical-testing}

We first give a classical deterministic algorithm for gradient testing. 

\begin{algorithm2e}[!ht]
\caption{Classical Deterministic Gradient Testing}
\label{algo:classical-testing}
\LinesNumbered
\DontPrintSemicolon

\KwInput{A point $\x\in\mathbb{R}^n$,  testing direction $\v$,   accuracy $\varepsilon\in(0,1/\sqrt{2})$, lower bound $\gamma$ on $\|\nabla f(\x)\|$ } 
\KwOutput{Test whether $\|\g-\v\|\le \varepsilon$ (YES case) or $\|\g - \v\|\ge 2\varepsilon$ (NO case)}\label{lin:classical testing output}

Set $\delta = \sqrt{1/(1-\varepsilon^2/2)^2 - 1}$, $\Delta_1 = \gamma/(7n)$, $\Delta_2 = \frac{\gamma}{8n^2}$, $\Delta_3 = \frac{\gamma\delta}{30\sqrt{14}n^{1.5}}$ and define $w_i(\beta) := \beta\mathbf{e}_1 - \mathbf{e}_i$\label{lin:testing-set}

For any $i=2,  3,  \ldots,  n$   initialize all $\ell_i=1$, $t_i = 1$

\For{$i=2$ \emph{to} $n$}{
Call \algo{dp} with input $(\x, \mathbf{e}_i, \Delta_1)$

\If{$\langle \nabla f, \mathbf{e}_i\rangle <\Delta_1$}{
$t_i \leftarrow -t_i$
}
}

Let $D = \diag(1, t_2,\dots,t_n)$ and apply the orthogonal transformation $D$ to the coordinate frame. 

Set $\y = (2n, -1,\dots,-1)$ and call \algo{dp} with input $(\x, \y/\|\y\|, \Delta_2)$

\If{$\langle \nabla f, \y/\|\y\|\rangle\le\Delta_2$}{\label{lin:basic-testing}
\Return No
}
 
\For{$i=2$ \emph{to} $n$}{

\RepeatBlock{\label{lin:begin-testing-multiply}
Set $\beta = \delta\ell_i/\sqrt{7n}$, and call \algo{dp} with input $\left(\x, w_i(\beta)/\|w_i(\beta)\| ,  \Delta_3\right)$ \label{lin:testing-call-dp}

\If{$ \langle \nabla f, w_i(\beta)/\|w_i(\beta)\|\rangle\le \Delta_3$}{\label{lin:recursive-judge}
$\ell_i \leftarrow 3\ell_i/2$
}
\Else{
\textbf{break}
}
\If{$\sum_{j=2}^n\ell_j^2\geq 21n$}{
\textbf{break} and \Return No
}
}\label{lin:end-testing-multiply}
}

\Return Yes

\end{algorithm2e}

\begin{theorem}\label{thm:classical-testing-deterministic}
\algo{classical-testing} solves \prob{testing} using $O(n)$ queries to a comparison oracle $O_f^{\mathrm{Comp}}$ \eqref{eq:comparison}.
\end{theorem}

Denote $\alpha_i=|g_i|/|g_1|$ for any $i=2,  3,  \ldots,  n$, where $\g$ is the normalized gradient (see \sec{notation}). Define $\delta = \sqrt{1/(1-\varepsilon^2/2)^2 - 1}$.  
The distance $\|\g-\v\|$ captures angular deviation, while the ratios $\alpha_i$ capture relative coordinate imbalance. The following lemma establishes that, to solve \prob{testing}, it suffices for us to estimate each $\alpha_i$ within a constant multiplicative error.

\begin{lemma}\label{lem:problem-reduce-main}
    If $\|\g - \v\|\le \varepsilon$, we have $\sum_{i=2}^n\alpha_i^2\le \delta^2$ and $g_1^2\sum_{i=2}^n\alpha_i^2\le \varepsilon^2$. If $\|\g - \v\|\ge 2\varepsilon$, we have $\sum_{i=2}^n\alpha_i^2\ge 4\delta^2$ and $g_1^2\sum_{i=2}^n\alpha_i^2\ge 2\varepsilon^2$. 
\end{lemma}

\begin{proof}
    Due to $\v = \mathbf{e}_1 = (1, 0, \dots, 0)$ and $\|\g\| = 1$, we have
    \begin{align*}
        \|\g - \v\|^2 = (1-g_1)^2 + g_2^2+ \dots + g_n^2=(1-g_1)^2+1-g_1^2=2-2g_1. 
    \end{align*}
    If $\|\g - \v\|\le \varepsilon$, we can get $g_1 \ge 1 - \varepsilon^2/2$, which means
    \[
    \sum_{i=2}^n\alpha_i^2 = (1-g_1^2)/g_1^2 \le 1/(1-\varepsilon^2/2)^2 - 1\le\delta^2. 
    \]
    If $\|\g - \v\|\ge 2\varepsilon$, we can get $g_1 \le 1 - 2\varepsilon^2$, which means
    \begin{align*}
        \sum_{i=2}^n\alpha_i^2 &= (1-g_1^2)/g_1^2 \ge 1/(1-2\varepsilon^2)^2 - 1
        \geq4\delta^2. 
    \end{align*}
\end{proof}

Intuitively, \algo{classical-testing} proceeds as follows: We first choose a suitable $\y$ to exclude the case when $\langle \v,\g\rangle$ is too small, which implies $\v$ belongs to the NO case. Then we call \algo{dp} and use $w_i(\beta) := \beta\mathbf{e}_1 - \mathbf{e}_i$ to compare $\beta$ and $|g_i/g_1|$ for each $i$. Note that if we call \algo{dp} with input direction $w_i(\beta)$, we can roughly get the sign of
\begin{align}\label{eq:sign-judgement}
    \langle \nabla f, w_i(\beta)/\|w_i(\beta)\| \rangle = \|\nabla f\|(\beta g_1 - g_i)/\|w_i(\beta)\|, 
\end{align}
which is determined by $\beta - g_i/g_1$. Therefore we can roughly get the sign of $\beta - g_i/g_1$ by it. 

If $\beta$ is roughly less than $|g_i/g_1|$, it is multiplied by $3/2$, and finally we find $\beta = \delta\ell_i/\sqrt{7n}$ which satisfies $2\beta/3<|g_i/g_1|<\beta$ approximately. And then we can count $\sum_{i=2}^n \ell_i^2$ to get the range of $\|\v - \g\|$.

\begin{proof}[Proof of \thm{classical-testing-deterministic}]

In \lin{basic-testing}, we exclude the extreme case where $g_1 = \langle \g,\v\rangle$ is very small. On the one hand, if
\[
\langle \nabla f, \y/\|\y\|\rangle\le \Delta, 
\]
we have
\[
2ng_1\le \sum_{i=2}^n g_i + \frac{4n\Delta_2}{\gamma}\le \sqrt{n\sum_{i=2}^n g_i^2}+\frac{1}{2n}=\sqrt{n(1-g_1^2)} + \frac{1}{2n}\le \sqrt{n}+\frac{1}{2n}, 
\]
which means
\[
\langle \v, \g\rangle =g_1\le \frac{1}{\sqrt{n}}.
\]
We can get that $\v$ belongs to the NO case. On the other hand, if
\[
\langle \nabla f, \y/\|\y\|\rangle\ge -\Delta,
\]
we have
\[
2ng_1\ge \sum_{i=2}^n g_i - \frac{4n\Delta_2}{\gamma}\ge \frac{1}{2}\sum_{i=2}^n g_i^2 - \frac{1}{2n} = \frac{1}{2}(1-g_1^2)-\frac{1}{2n},
\]
which means $g_1\ge 1/10n$. Therefore, in the rest of the proof, we assume
\begin{align}\label{eq:testing-assumption}
    g_1 = \langle \g, \v\rangle \ge \frac{1}{10n}.
\end{align}
With \eq{sign-judgement}, in our Lines~\ref{lin:begin-testing-multiply}--\ref{lin:end-testing-multiply}, if we get $\delta\ell_i/\sqrt{7n} - |\alpha_i|$ is negative within error $\Delta_3$, we will multiply $\ell_i$ by $3/2$, and if $\ell_i$ remains unchanged, we promise that we find the smallest integer $\ell_i$ which is an integer power of $3/2$ and satisfies $\delta\ell_i/\sqrt {7n}\geq |\alpha_i|$ (up to $\Delta_3$, the precision of calling \algo{dp} in \lin{testing-call-dp}). Consequently, for any $i=2,  3,  \ldots,   n$,   we can find such an $\ell_i$ using $\lceil1+\log\ell_i\rceil$ comparison queries. 

On the one hand, we prove that if $\sum_{j=2}^n \ell_j^2\ge 21n$, we will have $\|\g-\v\|\ge\varepsilon^2$. Denote $S=\{i\mid i\in\{2,  3,  \ldots,  n\},  \ell_{i}>1\}$.   Note that if $i\in S$,   it means that we have started the recursive setting of powers of $3/2$ in Lines~\ref{lin:begin-testing-multiply}--\ref{lin:end-testing-multiply} above, and we can finally get $\ell_i$ which do not satisfy the judge in \lin{recursive-judge}. It means that we have
\begin{align}
    \langle \nabla f, w_i(2\delta\ell_i/3\sqrt{7n})/\|w_i(2\delta\ell_i/3\sqrt{7n})\|\rangle\le \Delta_3, \label{eq:testing1}\\
    \langle \nabla f, w_i(\delta\ell_i/\sqrt{7n})/\|w_i(\delta\ell_i/\sqrt{7n})\|\rangle\ge -\Delta_3.\label{eq:testing2}
\end{align}
Combine \eq{sign-judgement} and \eq{testing1}, we can get
\begin{align*}
    \frac{2\delta\ell_i}{3\sqrt{7n}}g_1 - g_i \le \frac{\Delta_3}{\gamma}&\sqrt{1+\frac{4\delta^2\ell_i^2}{63n}} \le \sqrt{2}\Delta_3/\gamma,\\
    |\alpha_i| = |g_i/g_1|&\ge \frac{2\delta\ell_i}{3\sqrt{7n}}-\sqrt{2}\Delta_3/(\gamma g_1). 
\end{align*}
By our choice of $\Delta_3$ in Line~\ref{lin:testing-set} and \eq{testing-assumption}, we have
\[
|\alpha_i|\ge \frac{2\delta\ell_i}{3\sqrt{7n}}-\frac{\delta}{3\sqrt{7n}}.
\]
Applying the following Cauchy-Schwarz inequality,
\begin{align*}
    \left(|\alpha_i|^2 + \frac{\delta^2}{9n}\right)\cdot\left(1+\frac{1}{7}\right)\ge\left(|\alpha_i| + \frac{\delta}{3\sqrt{7n}}\right)^2,
\end{align*}
we obtain
\begin{align*}
    |\alpha_i|^2 &\ge \frac{\delta^2\ell_i^2}{18n}-\frac{\delta^2}{9n}.
\end{align*}
Therefore, if $\sum_{i=2}^n\ell_i^2\geq 21n$,   we have $\sum_{i\in S}\ell_i^2\geq 20n$,  and
\begin{align*}
    \sum_{i\in S} |\alpha_i|^2 \geq \sum_{i\in S} \left(\frac{\delta^2\ell_i^2}{18n}-\frac{\delta^2}{9n}\right)\geq \delta^2\left(\frac{20}{18}-\frac{1}{9}\right)=\delta^2. 
\end{align*}
Under the promise of \prob{testing} and \lem{problem-reduce-main}, this implies that $\x$ belongs to the NO case. 

On the other hand, for the case where $\sum_{i=2}^n\ell_i^2\leq 21n$. From \eq{testing2}, we have 
\[
|\alpha_i|\le \frac{\ell_i}{\sqrt{7n}} + \sqrt{1+\frac{\delta^2\ell_i^2}{7n}}\Delta_3/(\gamma g_1).
\]
Use the value of $\Delta_3$ we set in Line~\ref{lin:testing-set} and \eq{testing-assumption}, we have
\[
|\alpha_i|\le \frac{\delta\ell_i}{\sqrt{7n}} + \frac{\delta}{3\sqrt{7n}}.
\]
By Cauchy-Schwarz inequality, 
\begin{align*}
    |\alpha_i|^2\le \left(1+\frac{1}{6}\right)\left(\frac{\delta^2\ell_i^2}{7n} + \frac{\delta^2}{63n}\right) \le \frac{\delta^2\ell_i^2}{6n} + \frac{\delta^2}{9n}.
\end{align*}
If $\sum_{i=2}^n\ell_i^2\leq 21n$, we have
\begin{align*}
    \sum_{i=2}^n |\alpha_i|^2 \le \sum_{i=2}^n \left(\frac{\delta^2\ell_i^2}{6n} + \frac{\delta^2}{9n}\right)\leq \delta^2\left(\frac{21}{6}+\frac{1}{9}\right) < 4 \delta^2. 
\end{align*}
Under the promise of \prob{testing} and \lem{problem-reduce-main}, this implies that $\x$ belongs to the YES case.  

Next, we bound the query complexity of \algo{classical-testing}. Note that when \algo{classical-testing} terminates,  all $\ell_i$ satisfy
\begin{align*}
    \sum_{i=2}^n\ell_i \leq \sum_{i=2}^n\ell_i^2 \leq \frac{9}{4}\cdot \sum_{i=2}^n\max\{2\ell_i/3,  1\}^2=O(n). 
\end{align*}
Based on our recursive setting in Lines~\ref{lin:begin-testing-multiply}--\ref{lin:end-testing-multiply}, using 
\begin{align}
    \sum_{i=1}^n(1+\log \ell_i)\leq n+\sum_{i=1}^n\ell_i
\end{align}
queries, we can find all values of $\ell_i$ with $O(n)$ queries. 
\end{proof}

\subsection{Classical Lower Bound on Gradient Testing}\label{append:lower-classical-testing}
Here, we prove our classical deterministic lower bound for gradient testing, which implies that our \algo{classical-testing} is optimal. Our lower bound is obtained by considering the function $f(\x) = \langle\g, \x\rangle + b$ for some $\g\in\R^n$ and $b\in\R$, whose comparison query oracle satisfies
\begin{align}\label{eq:classical-reduce-oracle}
    O_f^{\comp}(\x, \x+\y) = \sgn(\langle\nabla f(\x),\y\rangle) = \sgn(\langle\g, \y\rangle). 
\end{align}

\begin{theorem}[Deterministic lower bound for gradient testing]
\label{thm:det-lb}
Fix $\varepsilon\in(0,1/10)$ and consider the objective
\[
f_{\g,b}(\x)=\langle \g,\x\rangle+b.
\]
Any deterministic algorithm that, for every $b\in\mathbb R$ and unit vectors $\g,\v\in\mathbb R^n$, correctly distinguishes
\[
\textnormal{(YES)}\quad \|\g-\v\|_2\le \varepsilon
\qquad\textnormal{from}\qquad
\textnormal{(NO)}\quad \|\g-\v\|_2>2\varepsilon
\]
using access to $O_{f_{\g,b}}^{\comp}$ must make at least $n-1$
comparison queries. 
\end{theorem}

\begin{proof}
By rotation invariance, assume $\v=\e_1$. Fix a deterministic algorithm $\mathcal A$ making at most $q\le n-2$ comparison queries. We construct two
affine objectives, one YES and one NO, whose comparison transcripts are
identical. Set $b=0$ and first consider the YES gradient
\[
\g_{\mathrm Y}=\e_1 .
\]
Run $\mathcal A$ with the oracle for
$f_{\mathrm Y}(\x)=\langle \g_{\mathrm Y},\x\rangle$, let
$\w^{(1)},\ldots,\w^{(q)}$ be the queries made on this execution path, and define
\[
W:=\operatorname{span}\{\w^{(1)},\ldots,\w^{(q)}\}.
\]
Since $\dim W\le q\le n-2$,
\[
\dim(\e_1^\perp\cap W^\perp)\ge n-1-q\ge 1 .
\]
Choose a unit vector $\u\in \e_1^\perp\cap W^\perp$. Now define the NO
gradient
\[
\g_{\mathrm N}:=\frac{\e_1+\delta \u}{\sqrt{1+\delta^2}},
\qquad
\delta:=3\varepsilon .
\]
Then $\|\g_{\mathrm N}\|=1$ and
\[
\|\g_{\mathrm N}-\e_1\|
\ge
|\langle \g_{\mathrm N}-\e_1,\u\rangle|
=
\frac{\delta}{\sqrt{1+\delta^2}}
>
2\varepsilon,
\]
where the last inequality uses $\delta=3\varepsilon$ and $\varepsilon<1/10$, indicating that $\g_{\mathrm N}$ is a NO
instance.

It remains to show that $\g_{\mathrm N}$ produces an identical transcript as $\g_{\mathrm Y}$. Proceed
by induction on the interaction. Suppose the transcripts agree up to query
$t-1$, for some $t< q$. Since $\mathcal A$ is deterministic, its next query
on $\g_{\mathrm N}$ is the same YES-path query $\w^{(t)}$. For this query,
since $\w^{(t)}\in W$ and $\u\perp W$,
\[
\langle \g_{\mathrm N},\w^{(t)}\rangle
=
\frac{\langle \e_1,\w^{(t)}\rangle+\delta\langle \u,\w^{(t)}\rangle}
{\sqrt{1+\delta^2}}
=
\frac{\langle \e_1,\w^{(t)}\rangle}{\sqrt{1+\delta^2}}.
\]
Hence
\[
\operatorname{sgn}(\langle \g_{\mathrm N},\w^{(t)}\rangle)
=
\operatorname{sgn}(\langle \e_1,\w^{(t)}\rangle)
=
\operatorname{sgn}(\langle \g_{\mathrm Y},\w^{(t)}\rangle).
\]
Thus the oracle replies agree at query $t$, completing the induction. After the first $q$ replies, $\mathcal A$ is in the same state as on the YES execution and therefore has same output. Since $\g_{\mathrm Y}$ is a YES instance and $\g_{\mathrm N}$ is a NO instance, $\mathcal A$ errs on one of them, and we can conclude that any correct deterministic algorithm needs at least $n-1$ queries.
\end{proof}

\section{Quantum Gradient Estimation}
\label{append:quantum-gradient-estimation}
Here, we prove the following quantum gradient estimation algorithm adapted from~\cite{Jordan_2005} estimates gradients to certain precision assuming an approximate quantum evaluation oracle:

\begin{proposition}[Quantum gradient estimation]
\label{prop:qft}
Let $n\ge 2$ and $t\in\mathbb{Z}_+$. Define $T:=t+1$ and
\[
M:=\Big\{0,\frac{1}{t},\frac{2}{t},\dots,1\Big\}^n,
\qquad N:=|M|=T^n .
\]
For any $\x\in[0,1]^n$, define the quantum state
\begin{equation}
\label{eq:Phi-x}
|\Phi_\x\rangle
:=\frac{1}{T^{n/2}}
\sum_{\y\in\{0,1,\dots,T-1\}^n}
e^{2\pi i \langle \y,\x\rangle}\,|\y\rangle .
\end{equation}
Let $F_T$ denote the $T$-dimensional quantum Fourier transform over $\mathbb{Z}_T$, i.e.,  $F_T|j\rangle = \frac{1}{\sqrt{T}}\sum_{k = 0}^{T-1}\e^{-2\pi i jk/T}|k\rangle$, and denote $F:=F_T^{\otimes n}$. Consider the following procedure: apply $F^\dagger$ to the input state, which is $|\Phi_\x\rangle$ in ideal case and $|\psi\rangle$ in approximation case, measure in
the computational basis obtaining $K\in\{0,1,\dots,T-1\}^n$, and output $\hat \v:=\frac{1}{t}K\in M$. Let
\begin{equation}
\label{eq:m-choice}
m:=\left\lceil 2+\frac{3}{2}n\right\rceil .
\end{equation}
Then for every $\x\in[0,1]^n$,
\begin{equation}
\label{eq:success-main}
\Pr\!\left[\ \|\hat \v-\x\|_2 \le \frac{\sqrt{n}\,(m+1)}{t}\ \right]\ge \frac{2}{3}.
\end{equation}
Moreover, if the actual input state $|\psi\rangle$ satisfies $\|\,|\psi\rangle-|\Phi_\x\rangle\,\|_2 \le \varepsilon$, then the output satisfies 
\begin{equation}
\label{eq:robust-main}
\Pr\left[\ \|\hat \v-\x\|_2 \le \frac{\sqrt{n}\,(m+1)}{t}\ \right]
\ge \frac{2}{3}-2\varepsilon .
\end{equation}
\end{proposition}

\begin{proof}
We first analyze the ideal input $|\Phi_\x\rangle$. Since
\[
e^{2\pi i\langle \y,\x\rangle}=\prod_{j=1}^n e^{2\pi i y_j x_j}
\quad\text{and}\quad
F=F_T^{\otimes n},
\]
the state $|\Phi_\x\rangle$ decomposes as a tensor product across coordinates.
Consequently, after applying $F^\dagger$, the measurement outcome
$K=(K_1,\dots,K_n)$ consists of independent coordinates, where each $K_j$ arises
from measuring
\[
|\phi_{\alpha}\rangle
:=\frac{1}{\sqrt{T}}\sum_{y=0}^{T-1} e^{2\pi i y\alpha}\,|y\rangle,
\qquad \alpha=x_j .
\]
Applying $F_T^\dagger$ to $|\phi_\alpha\rangle$, the amplitude of observing
$k\in\{0,\dots,T-1\}$ is
\[
a_k
=\langle k|F_T^\dagger|\phi_\alpha\rangle
=\frac{1}{T}\sum_{y=0}^{T-1} e^{2\pi i y(\alpha-k/T)} .
\]
Let $\delta_k:=\alpha-k/T$. Summing the geometric series gives
\[
a_k
=\frac{1}{T}\cdot
\frac{1-e^{2\pi i T\delta_k}}{1-e^{2\pi i\delta_k}}
=\frac{1}{T}\,
e^{\pi i (T-1)\delta_k}
\frac{\sin(\pi T\delta_k)}{\sin(\pi\delta_k)} .
\]
Hence
\begin{align*}
\Pr[K_j=k]
=\frac{1}{T^2}
\left(\frac{\sin(\pi T\delta_k)}{\sin(\pi\delta_k)}\right)^2 .
\end{align*}
Let $\theta:=T\alpha$. Then $\delta_k=(\theta-k)/T$ and
\[
\Pr[K_j=k]
=\frac{1}{T^2}
\left(\frac{\sin(\pi(\theta-k))}{\sin(\pi(\theta-k)/T)}\right)^2 .
\]
Using $|\sin(\pi z)|\le 1$ and the inequality
\[
|\sin(\pi u)|\ge 2|u| \quad \text{for } |u|\le \tfrac12,
\]
we obtain, for all $k$,
\[
\Pr[K_j=k]\le \frac{1}{4(\theta-k)^2}.
\]
Therefore, for any integer $m\ge 2$,
\begin{equation}
\label{eq:1d-tail}
\Pr\big[|K_j-\theta|\ge m\big]
\le \sum_{r=m}^{\infty}\frac{1}{2r^2}
\le \frac{1}{2(m-1)} .
\end{equation}
Let $\theta_j:=T x_j$. By \eq{1d-tail},
\[
\Pr\big[|K_j-\theta_j|\ge m\big]\le \frac{1}{2(m-1)} .
\]
Applying the union bound,
\[
\Pr\big[\exists j:\ |K_j-\theta_j|\ge m\big]
\le \frac{n}{2(m-1)} .
\]
With the choice of \eq{m-choice}, this probability is at most $1/3$, hence
\[
\Pr\big[\forall j:\ |K_j-\theta_j|<m\big]\ge \frac{2}{3}.
\]
On this event,
\[
|\hat v_j-x_j|
\le \frac{m+1}{t},
\]
and thus
\[
\|\hat \v-\x\|_2
\le \frac{\sqrt{n}\,(m+1)}{t}.
\]
This proves \eq{success-main}. Since $F^\dagger$ is unitary,
\[
\big\|\,F^\dagger|\psi\rangle-F^\dagger|\Phi_\x\rangle\,\big\|_2
=\big\|\,|\psi\rangle-|\Phi_\x\rangle\,\big\|_2
\le \varepsilon .
\]
Let $P$ and $Q$ denote the measurement distributions obtained from
$F^\dagger|\psi\rangle$ and $F^\dagger|\Phi_\x\rangle$, respectively. For any event $\textsf E$,
\[
|P(\textsf E)-Q(\textsf E)|
\le \tfrac12\big\|\,|\psi\rangle\langle\psi|
-|\Phi_\x\rangle\langle\Phi_\x|\,\big\|_1
\le \big\|\,|\psi\rangle-|\Phi_\x\rangle\,\big\|_2
\le \varepsilon .
\]
Let $\textsf E$ be the event
\[
\|\hat \v-\x\|_2 \le \frac{\sqrt{n}\,(m+1)}{t}.
\]
Since $Q(\textsf E)\ge 2/3$, it follows that
\[
P(\textsf E)\ge \frac{2}{3}-\varepsilon \ge \frac{2}{3}-2\varepsilon,
\]
which proves \eq{robust-main}.
\end{proof}

\section{Proof Details of Gradient Estimation Lower Bounds}\label{append:lower}

\subsection{Classical Lower Bound}\label{append:lower-classical-estimation}

In this subsection, we prove the classical lower bound on gradient estimation (\thm{classical-estimation-lower}). Similarly, this is also obtained by considering to the special case in \append{lower-classical-testing} and our query oracle becomes \eq{classical-reduce-oracle}. Our lower bound relies on the following two lemmas.

\begin{lemma}\label{lem:separated-set}
Let $n\ge 1$ and $0<\varepsilon\le 1$.  
There exists a set $M\subset S_n$ such that
\[
\forall \x\neq \y\in M,\quad \|\x-\y\|>\varepsilon,
\quad\text{and}\quad
|M| \ge \frac{2}{\varepsilon^{n}}.
\]
\end{lemma}

\begin{proof}
We construct $M$ greedily as a \emph{maximal} $\varepsilon$-separated subset of $S_n$:
start with $M=\emptyset$ and keep adding points $\x\in S_n$ with $\|\x-\y\|>\varepsilon$ for all
$\y\in M$, until this is no longer possible. By construction, $M$ is $\varepsilon$-separated. Maximality implies the covering property
\[
S_n \subseteq \bigcup_{\x\in M} B(\x,\varepsilon),
\qquad
B(\x,\varepsilon):=\{\z\in S^n:\|\z-\x\|\le \varepsilon\},
\]
because otherwise a point outside the union could be added to $M$, contradiction. Hence, writing $\sigma_n(\cdot)$ for the standard rotation-invariant surface measure on $S_n$, and equivalently, $\sigma_n(A)/\sigma_n(S_n)$ is the probability that a uniformly random point on $S_n$ lies in $A$. Then we have
\[
\sigma_n(S_n)\ \le\ \sum_{\x\in M} \sigma_n(B(\x,\varepsilon))
\ \le\ |M|\,\sup_{\x\in S_n}\sigma_n(B(\x,\varepsilon)).
\]
So it suffices to show the following: for every $\x\in S_n$ and $0<\varepsilon\le 1$,
\begin{equation}\label{eq:cap-bound}
\sigma_n(B(\x,\varepsilon)) \ \le\ \frac{1}{2}\,\varepsilon^{\,n}\,\sigma_n(S_n).
\end{equation}
Assuming \eq{cap-bound} is true, we get
\[
\sigma_n(S_n)\ \le\ |M|\cdot \frac{1}{2}\varepsilon^n\,\sigma_n(S_n)
\quad\Rightarrow\quad
|M|\ \ge\ \frac{2}{\varepsilon^n},
\]
as claimed. In the rest of the proof, we prove \eq{cap-bound}. Fix $\x\in S_n$ and let $Z$ be uniform on $S_n$. Then
\[
\frac{\sigma_n(B(\x,\varepsilon))}{\sigma_n(S^n)} \ =\ \Pr\big[\|Z-\x\|\le \varepsilon\big].
\]
Let $T:=\langle Z,\x\rangle\in[-1,1]$. Since $\|Z-\x\|^2=2-2\langle Z,\x\rangle$, the event
$\|Z-\x\|\le \varepsilon$ is equivalent to
\[
T \ \ge\ 1-\frac{\varepsilon^2}{2}.
\]
The density of $T$ is
\[
f_n(t)=c_n(1-t^2)^{\frac{n-1}{2}},\qquad
c_n:=\frac{\Gamma\!\left(\frac{n+1}{2}\right)}{\sqrt{\pi}\,\Gamma\!\left(\frac{n}{2}\right)}.
\]
For $t\in[0,1]$ we have $1-t^2=(1-t)(1+t)\le 2(1-t)$, so with $\delta:=\varepsilon^2/2$,
\begin{align*}
\Pr\Big[T\ge 1-\delta\Big]
&= \int_{1-\delta}^{1} c_n(1-t^2)^{\frac{n-1}{2}}\,\d t \\
&\le \int_{1-\delta}^{1} c_n\big(2(1-t)\big)^{\frac{n-1}{2}}\,\d t \\
&= c_n\,2^{\frac{n-1}{2}}\int_{0}^{\delta} u^{\frac{n-1}{2}}\,\d u
\qquad (u=1-t)\\
&= c_n\,2^{\frac{n-1}{2}}\cdot \frac{\delta^{\frac{n+1}{2}}}{\frac{n+1}{2}}
= \frac{c_n}{n+1}\,\varepsilon^{n+1}.
\end{align*}
Using the standard gamma-ratio bound
\begin{align*}
    c_n=\frac{\Gamma\left(\frac{n+1}{2}\right)}{\sqrt{\pi}\,\Gamma\left(\frac{n}{2}\right)}\le\ \sqrt{\frac{n}{2}}
\end{align*}
we obtain
\[
\Pr\big[\|Z-x\|\le \varepsilon\big]
=\Pr\Big[T\ge 1-\frac{\varepsilon^2}{2}\Big]
\le \frac{1}{2}\varepsilon^{n+1}
\le \frac{1}{2}\varepsilon^{n}
\]
for any $\varepsilon\leq 1$. This finishes the proof.
\end{proof}

We need another lemma on adaptive query lower bounds. We prove that any query model with three possible outcomes, corresponding to $<$, $>$, $=$ in our comparison query model, need to use at least $\log n$ queries to distinguish $n$ different results, no matter it is deterministic algorithm or random algorithm.
\begin{lemma}\label{lem:information-lower-bound}
Let $I$ be an unknown index uniformly distributed over $[n] := \{1,2,\dots,n\}$. Consider a query model in which each query returns one of three possible outcomes, and each $I$ corresponds to a type of query oracle. For different $I$, their oracles are also different. Any (possibly randomized) algorithm that identifies $I$ with success probability at least $2/3$ requires $\Omega(\log n)$ queries in the worst case.
\end{lemma}

\begin{proof}
Consider the uniform distribution over $I$. By Yao's minimax principle, there exists a fixing of the internal randomness of $\mathcal{A}$ such that the resulting deterministic algorithm $\mathcal{A}_r$ succeeds with probability at least $2/3$ when $I$ is drawn uniformly from $[n]$. Therefore, it suffices to prove that any deterministic algorithm that makes at most $T$ queries has success probability at most $O(3^T / n)$ under the uniform distribution. 

Now consider an arbitrary deterministic algorithm making at most $T$ adaptive queries. For such a determined algorithm, it can be represented as a ternary decision tree of depth at most $T$.
Each internal node corresponds to a query, and each node has exactly three outgoing edges, one for each possible query outcome.
Each leaf of the tree outputs a value in $[n]$.

A decision tree of depth at most $T$ has at most $3^T$ leaves, and hence the algorithm can correctly identify at most $3^T$ values of $I$. Under the uniform distribution over $I$, the success probability of any such deterministic algorithm is therefore at most $3^T / n$. To ensure that the success probability is least $2/3$, we must have $T \ge \log_3 n - O(1)$. This proves that any randomized algorithm identifying $I$ with constant success probability requires $\Omega(\log n)$ queries.
\end{proof}

\begin{proof}[Proof of \thm{classical-estimation-lower}]
    By \lem{separated-set}, we can get an $\varepsilon$-net $M$ on $S_n$, which satisfies that for each $\x,\y\in M$, we have $\|\x-\y\|>\varepsilon$ and $|M| = \Omega((1/\varepsilon)^n)$.

    Assume our gradient direction belongs to $M$. Note that each comparison query only has three types of possible responses. As a result, by \lem{information-lower-bound}, if we want to distinguish each gradient direction in $M$, the query complexity is at least
    \[
    \log_3|M| = \Omega(n\log(1/\varepsilon)).
    \]
\end{proof}

\subsection{Quantum Lower Bound}\label{append:lower-quantum-estimation}
In this subsection, we prove the quantum lower bound on gradient estimation (\thm{quantum-estimation-lower}). It is still obtained by considering to the special case in \append{lower-classical-testing} and our quantum oracle becomes
\begin{align}\label{quantum-reduce-oracle}
    O_{f,Q}^{\comp}|\x\rangle|\x+\y\rangle|z\rangle = |\x\rangle|\y\rangle|z\oplus\sgn(\langle\nabla f(\x),\y\rangle)\rangle.
\end{align}
Our proof is decomposed into three parts: a bound on average sensitivity of halfspaces (\append{average-sensitivity}), a bound on oracle distinction (\append{distinction}), and finally the proof of \thm{quantum-estimation-lower} (\append{quantum-lower-final}).

\subsubsection{Average sensitivity of halfspaces}\label{append:average-sensitivity}
\begin{lemma}[Average $k$-flip sensitivity of halfspaces]
\label{lem:average-sensitive}
There exists a universal constant $C'>0$ such that the following holds.  
Let
\[
h(\z)=\mathbf{1}\!\left[\sum_{i=1}^n w_i z_i \ge 0\right],  \qquad \z\in\{-1,  1\}^n,  
\]
be a halfspace.   Let $Z$ be uniform on $\{-1,  1\}^n$,   and let $S$ be uniform over all
$k$-subsets of $[n]$.   Then
\[
\mathbb{E}_{S:\,  |S|=k}\,  \Pr\big[h(Z)\neq h(Z^{\oplus S})\big]\ \le\ C'\sqrt{\frac{k}{n}}.  
\]
Moreover,   for all $k\le n/2$ one may take $C'$ universal (independent of $n,  k$).  
\end{lemma}

\begin{proof}

Scaling $\w$ by any $\lambda>0$ does not change $h$,   hence we may assume
\[
\|\w\|_2^2=\sum_{i=1}^n w_i^2=1.  
\]
Also,   flipping the sign of a coordinate $z_i$ and simultaneously flipping the sign of $w_i$
leaves the distribution of $Z$ invariant and does not change the probability of disagreement.  
Thus,   without loss of generality we may assume $w_i\ge 0$ for all $i$. Fix $S\subseteq[n]$ with $|S|=k$.   Write
\[
X :=\ \sum_{i=1}^n w_i Z_i,  
\qquad
B_S :=\ \sum_{i\in S} w_i Z_i,  
\qquad
Y_S :=\ \sum_{i\notin S} w_i Z_i.  
\]
Then $X=Y_S+B_S$.   Flipping the coordinates in $S$ sends $Z_i\mapsto -Z_i$ for $i\in S$,  
hence the corresponding linear form becomes
\[
X' \ :=\ \sum_{i=1}^n w_i (Z^{\oplus S})_i \ =\ Y_S - B_S.  
\]
Therefore,  
\[
h(Z)\neq h(Z^{\oplus S})
\quad\Longleftrightarrow\quad
\mathbf{1}[X\ge 0]\neq \mathbf{1}[X'\ge 0]
\quad\Longleftrightarrow\quad
(X)(X')<0\ \text{ or }\ XX'=0.  
\]
Since $XX'=(Y_S+B_S)(Y_S-B_S)=Y_S^2-B_S^2$,   we obtain the inclusion
\begin{equation}\label{eq:disagree-implies}
\{h(Z)\neq h(Z^{\oplus S})\}\ \subseteq\ \{|Y_S|\le |B_S|\}.  
\end{equation} 
Let
\[
\sigma_S^2\ :=\ \mathrm{Var}(Y_S)\ =\ \sum_{i\notin S} w_i^2 \ =\ 1-\sum_{i\in S} w_i^2
\ =\ 1-\|w_S\|_2^2.  
\]
Consider the normalized sum $\widetilde{Y}_S:=Y_S/\sigma_S$,   which has variance $1$.  
A standard anti-concentration inequality for Rademacher sums (a consequence of a local limit
theorem, see e.g.,~\citep{DDFS14}) states that there exists a universal constant $C_0>0$ such that
for every $t>0$,  
\begin{equation}\label{eq:anti-conc}
\Pr\big(|\widetilde{Y}_S|\le t\big)\ \le\ C_0\,  t.  
\end{equation}
Equivalently,   for every $u>0$,  
\begin{equation}\label{eq:anti-conc-unscaled}
\Pr\big(|Y_S|\le u\big)\ \le\ C_0\,  \frac{u}{\sigma_S}.  
\end{equation}
Since the random variables $Y_S$ and $B_S$ depend on disjoint sets of coordinates of $Z$, we can get that $Y_S$ and $B_S$ are independent. 
Using \eq{disagree-implies}, independence, and conditioning on $B_S$,  we have
\begin{align*}
\Pr\big(h(Z)\neq h(Z^{\oplus S})\big)
&\le \Pr\big(|Y_S|\le |B_S|\big)\\
&= \mathbb{E}\Big[\Pr\big(|Y_S|\le |B_S|\ \big|\ B_S\big)\Big]\\
&\le \mathbb{E}\Big[C_0\,  \frac{|B_S|}{\sigma_S}\Big]
\qquad\text{(by \eq{anti-conc-unscaled} with $u=|B_S|$)}\\
&= \frac{C_0}{\sigma_S}\,  \mathbb{E}|B_S|.  
\end{align*}
By Cauchy-Schwarz inequality, there exists a universal constant $C_1>0$ such that
\[
\mathbb{E}|B_S|
=\mathbb{E}\left|\sum_{i\in S} w_i Z_i\right|
\le C_1\left(\sum_{i\in S} w_i^2\right)^{1/2}
= C_1\,  \|w_S\|_2.  
\]
Hence
\begin{equation}\label{eq:fixed-S-bound}
\Pr\big(h(Z)\neq h(Z^{\oplus S})\big)\ \le\ C_0C_1\,  \frac{\|w_S\|_2}{\sqrt{1-\|w_S\|_2^2}}.  
\end{equation}
Define $a:=\|w_S\|_2\in[0,  1]$.   We claim there is a universal $C_2>0$ such that for all $a\in[0,  1]$,  
\begin{equation}\label{eq:denom-remove}
\min\left\{1,  \ \frac{a}{\sqrt{1-a^2}}\right\}\ \le\ C_2\,  a.  
\end{equation}
Indeed,   if $a\le 1/2$ then $\sqrt{1-a^2}\ge \sqrt{3}/2$ and
$\frac{a}{\sqrt{1-a^2}}\le \frac{2}{\sqrt{3}}a$.  
If $a>1/2$,   then $\min\{1,  \frac{a}{\sqrt{1-a^2}}\}\le 1 \le 2a$.  
Thus \eq{denom-remove} holds with $C_2:=2$.  

Combining \eq{fixed-S-bound} with the trivial bound
$\Pr(h(Z)\neq h(Z^{\oplus S}))\le 1$ and using \eq{denom-remove},   we get
a universal constant $C_3>0$ such that for every fixed $S$,  
\begin{equation}\label{eq:fixed-S-linear}
\Pr\big(h(Z)\neq h(Z^{\oplus S})\big)\ \le\ C_3\,  \|w_S\|_2.  
\end{equation}
Take the expectation over uniform $S$ with $|S|=k$:
\[
\mathbb{E}_{S}\Pr\big(h(Z)\neq h(Z^{\oplus S})\big)
\ \le\ C_3\,  \mathbb{E}_{S}\|w_S\|_2.  
\]
By Jensen's inequality,  
\[
\mathbb{E}_S\|w_S\|_2\ \le\ \sqrt{\mathbb{E}_S\|w_S\|_2^2}.  
\]
Moreover, 
\[
\mathbb{E}_S\|w_S\|_2^2
=\mathbb{E}_S\sum_{i\in S} w_i^2
=\sum_{i=1}^n w_i^2\,  \Pr(i\in S)
=\sum_{i=1}^n w_i^2\,  \frac{k}{n}
=\frac{k}{n}\|w\|_2^2
=\frac{k}{n}.  
\]
Therefore,  
\[
\mathbb{E}_{S:\,  |S|=k}\Pr\big[h(Z)\neq h(Z^{\oplus S})\big]
\ \le\ C_3\,  \sqrt{\frac{k}{n}}.  
\]
Setting $C':=C_3$ completes the proof.  
\end{proof}

\subsubsection{Distinction bound}\label{append:distinction}
\begin{lemma}\label{lem:original-bound}
Let $\z \in \{-1,  1\}^n$ and consider the oracle
\[
  O_\z \,  :\,   |\x,  \y\rangle \mapsto |\x,  \,   \y \oplus f(\x,  \z)\rangle,  
\]
where $\x \in \mathbb{Z}^n$ (coordinates not necessarily $\pm 1$) and
\[
  f(\x,  \z) =
  \begin{cases}
    1 & \text{if } \langle \x,  \z\rangle \ge 0,  \\
    0 & \text{if } \langle \x,  \z\rangle < 0.  
  \end{cases}
\]
Let
\[
  M := \{(\z,  \tilde \z) \in \{-1,  1\}^n \times \{-1,  1\}^n : H(\z,  \tilde \z) = k\},  
\]
where $H$ denotes the Hamming distance. Then the value
\[
  S_1 := \sum_{(\z,  \tilde \z)\in M} \bigl\|O_\z|\psi\rangle - O_{\tilde \z}|\psi\rangle\bigr\|^2
\]
for a normalized state $|\psi\rangle$ satisfies
\[
  S_1 = O\left(\sqrt{\frac{k}{n}} \binom{n}{k}\, 2^n\right).  
\]
\end{lemma}

\begin{proof}

Let the Hilbert space be spanned by $\{|\x,  \y\rangle\}$, where $\x$ is the query register and $\y$ is the answer register. Any normalized state can be written as
\[
  |\psi\rangle \;=\; \sum_{\x} |\x\rangle|\psi_\x\rangle,  
  \qquad \sum_\x \|\ket{\psi_\x}\|^2 = 1,  
\]
where $|\psi_\x\rangle$ is the (unnormalized) state on the second register conditional on query $\x$.  

For each fixed $\x$ and $\z$,   define the unitary on the answer register
\[
  U_\z^{(\x)} : |\y\rangle \mapsto |\y \oplus f(\x,  \z)\rangle.  
\]
Then
\[
  O_\z|\psi\rangle - O_{\tilde \z}|\psi\rangle
  \;=\;
  \sum_\x |\x\rangle\bigl(U_\z^{(\x)} - U_{\tilde \z}^{(\x)}\bigr)|\psi_\x\rangle.  
\]
Let
\[
  D(\z,  \tilde \z) \;:=\; \{\x : f(\x,  \z) \neq f(\x,  \tilde \z)\}.  
\]
If $\x\notin D(\z,  \tilde \z)$ then $U_\z^{(\x)}=U_{\tilde \z}^{(\x)}$ and that term vanishes.   Therefore
\begin{align*}
\bigl\|O_\z|\psi\rangle - O_{\tilde \z}|\psi\rangle\bigr\|^2
= \bigg\|\sum_{\x\in D(\z,  \tilde \z)}
|\x\rangle\bigl(U_\z^{(\x)} - U_{\tilde \z}^{(\x)}\bigr)|\psi_\x\rangle\bigg\|^2
= \sum_{\x\in D(\z,  \tilde \z)}
\bigl\|\bigl(U_\z^{(\x)} - U_{\tilde \z}^{(\x)}\bigr)|\psi_\x\rangle\bigr\|^2,  
\end{align*}
Now, for any two unitaries $U,V$ and any vector $|\v\rangle$,  
\[
  \|(U-V)|\v\rangle\|
  \;\le\; \|U|\v\rangle\| + \|V|\v\rangle\|
  \;=\; 2\|\v\|.  
\]
Hence,  
\[
  \bigl\|\bigl(U_\z^{(\x)} - U_{\tilde \z}^{(\x)}\bigr)|\psi_\x\rangle\bigr\|^2
  \;\le\; 4 \||\psi_\x\rangle\|^2,  
  \quad \text{whenever } f(\x,  \z)\neq f(\x,  \tilde \z).  
\]
We conclude that for any $\z,  \tilde \z$,  
\begin{align}
  \bigl\|O_\z|\psi\rangle - O_{\tilde \z}|\psi\rangle\bigr\|^2
  \;\le\; 4\sum_{\x : f(\x,  \z)\neq f(\x,  \tilde \z)} \||\psi_\x\rangle\|^2. \label{eq:distinction-basic}
\end{align}
Using \eq{distinction-basic},  
\begin{align*}
  S_1
  &\le
   4\sum_{(\z,  \tilde \z)\in M}
     \sum_{\x : f(\x,  \z)\neq f(\x,  \tilde \z)} \||\psi_\x\rangle\|^2 \\
  &= 4\sum_\x \||\psi_\x\rangle\|^2
       \underbrace{\bigl|\{(\z,  \tilde \z)\in M : f(\x,  \z)\neq f(\x,  \tilde \z)\}\bigr|}_{=: N_\x}.  
\end{align*}
Therefore,
\[
  S_1 \;\le\; 4\sum_x \||\psi_\x\rangle\|^2 N_\x
  \;\le\; 4\Bigl(\max_\x N_\x\Bigr)\sum_\x \||\psi_\x\rangle\|^2
  \;=\; 4\max_\x N_\x,  
\]
using $\sum_\x \|\ket{\psi_\x}\|^2 = 1$. Hence, the problem reduces to bounding $N_\x$, where for fixed $\x\in\mathbb{Z}^n$ we define
\[
  g_\x(\z) := f(\x,  \z) = \mathbf{1}[\langle \x,  \z\rangle \ge 0],  
  \quad \z\in\{-1,  1\}^n,  
\]
and
\[
  N_\x = |\{(\z,  \tilde \z)\in M : g_\x(\z)\neq g_\x(\tilde \z)\}|.  
\]
Note that the function $g_\x$ is a \emph{halfspace} (linear threshold function) on the hypercube:
\[
  g_\x(\z) = \mathbf{1}\biggl[\sum_{i=1}^n w_i z_i \ge 0\biggr],  
  \qquad w_i := x_i \in \mathbb{Z}.  
\]
Applying \lem{average-sensitive} to $g_\x$, we obtain
\[
  N_\x = \binom{n}{k} 2^{n-1} \mathbb{E}_{S:\,  |S|=k}\,  \Pr\big[g_\x(Z)\neq g_\x(Z^{\oplus S})\big]
  \le C_0 \sqrt{\frac{k}{n}}\, \binom{n}{k} 2^{n-1}.  
\]
Recalling that
\[
  S_1 \le 4 \max_x N_\x,  
\]
we conclude
\[
  S_1 \;\le\; 4 \cdot C_0 \sqrt{\frac{k}{n}}\,\binom{n}{k}2^{n-1}
  = 2C_0\,   \sqrt{\frac{k}{n}}\,  \binom{n}{k}2^{n}.  
\]
Thus there is an absolute constant $C := 2C_0 > 0$ such that
\[
  S_1 \le C\sqrt{\frac{k}{n}}\, \binom{n}{k}2^{n}=O\left(\sqrt{\frac{k}{n}}\, \binom{n}{k}2^{n}\right).
\]
\end{proof}

\subsubsection{Proof of \thm{quantum-estimation-lower}}\label{append:quantum-lower-final}
Now, we give the proof of \thm{quantum-estimation-lower}. This is established by the following theorem:
\begin{theorem}\label{thm:lower bound}
Let $\z \in \{-1,  1\}^{n}$ be an unknown string and we have oracle $O_\z |\x,  \y\rangle = |\x,  \y\oplus f(\x,  \z)\rangle$,  where $\x \in \Z^n$ and $f(\x,  \z)=1$ when $\langle \x,  \z \rangle \ge 0$,   $f(\x,  \z)=0$ when $\langle \x,  \z \rangle < 0$.   Then every algorithm output $\z$ within $k$ errors uses at least $\frac{1}{4} \log \frac{n}{k}$ queries to $O_\z$.  
\end{theorem}

\begin{proof}
   Define $M=\{(\x,  \tilde \x)\mid H(\x,  \tilde \x)=k\}$ and we can get $|M|=\binom{n}{k}\cdot2^n$.  
   By \lem{original-bound},  we have
   \begin{align*}
       S_1:=\sum_{(\x,  \tilde \x) \in M}\lVert O_x|\psi\rangle-O_{\tilde \x}|\psi\rangle \rVert^{2}\le C\sqrt{\frac{k}{n}}\, \binom{n}{k}2^{n}.
   \end{align*}
   We hope to get a bound of
   \begin{align*}
       S_T:=\sum_{(\x,  \tilde \x) \in M}\lVert O_\x U_{T-1}O_\x\cdots U_1O_\x|\psi\rangle-O_{\tilde \x}U_{T-1}O_{\tilde \x}\cdots U_1O_{\tilde \x}|\psi\rangle \rVert^{2}.
   \end{align*}
   In fact,  we have
   \begin{align*}
       \sum_{(\x,  \tilde \x) \in M}\lVert O_\x U_1O_\x|\psi\rangle-O_{\tilde \x}U_1O_{\tilde \x}|\psi\rangle \rVert^{2}
       \le2\sum_{(\x,  \tilde \x) \in M}\lVert (O_\x-O_{\tilde \x})U_1O_x|\psi\rangle \rVert^{2}+\lVert O_\x|\psi\rangle-O_{\tilde \x}|\psi\rangle \rVert^{2}.
   \end{align*}
   Assume that $|\psi\rangle = \sum_i\alpha_i|\psi_i\rangle|\phi_i\rangle$.  
   Since our oracle can only judge,  which means for different $\x$,  $O_\x|\psi_i\rangle|\phi_i\rangle$ only have two results $|\psi_i\rangle|\phi_{i,  0}\rangle$ and $|\psi_i\rangle|\phi_{i,  1}\rangle$.   So we can get
   \begin{align*}
       \sum_{(\x,  \tilde \x) \in M}\lVert (O_\x-O_{\tilde \x})U_1O_x|\psi\rangle \rVert^{2}
       &\le \sum_{(\x,  \tilde \x) \in M} \sum_i \alpha_i^2\lVert (O_\x-O_{\tilde \x})U_1O_\x|\psi_i\rangle|\phi_i\rangle \rVert^{2}\\
       &\le \sum_{(\x,  \tilde \x) \in M} \sum_i \alpha_i^2(\lVert (O_\x-O_{\tilde x})U_1|\psi_i\rangle|\phi_{i,  0}\rangle \rVert^{2}+\lVert (O_\x-O_{\tilde x})U_1|\psi_i\rangle|\phi_{i,  1}\rangle \rVert^{2})\\
       &=\sum_i \alpha_i^2\sum_{(\x,  \tilde \x) \in M} \lVert (O_\x-O_{\tilde \x})U_1|\psi_i\rangle|\phi_{i,  0}\rangle \rVert^{2}+\lVert (O_\x-O_{\tilde \x})U_1|\psi_i\rangle|\phi_{i,  1}\rangle \rVert^{2}\\
       &\le \sum_i 2\alpha_i^2S_1 = 2S_1.
   \end{align*}
   Similarly,   we can get
   \begin{align*}
       S_T\le (2^{T+1}+2)S_1\le (2^{T+1}+2)\cdot \sqrt{\frac{k}{n}}\, \binom{n}{k}2^{n}.
   \end{align*}
   We let $T=\frac{1}{4}\log \frac{n}{k}$,   and then there exist a $(\x,  \tilde \x) \in M$ such that
   \begin{align*}
       \lVert O_\x U_{T-1}O_\x\cdots U_1O_\x|\psi\rangle-O_{\tilde \x}U_{T-1}O_{\tilde \x}\cdots U_1O_{\tilde \x}|\psi\rangle \rVert^{2}\le \frac{1}{(n/k)^{\frac{1}{4}}}.  
   \end{align*}
\end{proof}

Note that we can reduce \prob{estimation} to the special case of assuming $\nabla f \in \{-1,  1\}^n$, and the result allows $\varepsilon n$-bit error. By \thm{lower bound},  set $k = \varepsilon n$, we obtain the quantum lower bound $\Omega(\log (1/\varepsilon))$ for \thm{quantum-estimation-lower}.  
\end{document}